\pdfoutput=1

\documentclass[11pt]{article}

\usepackage[final]{acl}

\usepackage{CJK}
\usepackage{times}
\usepackage{latexsym}
\usepackage{graphicx}
\usepackage{subcaption}
\usepackage{multirow}
\usepackage{booktabs}
\usepackage[table]{xcolor}  
\usepackage{colortbl}  
\usepackage[most]{tcolorbox}
\usepackage{amsmath, amssymb}
\usepackage{tabularx}
\usepackage{natbib}
\usepackage{xcolor}

\usepackage[noabbrev,capitalize,nameinlink]{cleveref}

\definecolor{darkorange}{rgb}{0.75,0.34,0.0}  
\definecolor{lightyellow}{rgb}{1.0,1.0,0.59}  

\usepackage[T1]{fontenc}

\usepackage[utf8]{inputenc}

\usepackage{microtype}

\usepackage{inconsolata}

\usepackage{graphicx}

\usepackage{amsmath}
\usepackage{amssymb}

\usepackage{mathtools}
\usepackage{amsthm}
\theoremstyle{plain}
\newtheorem{theorem}{Theorem}[section]

\theoremstyle{definition}
\newtheorem{definition}[theorem]{Definition}

\theoremstyle{remark}

\usepackage{siunitx}

\usepackage{arydshln}

\title{Calibration Is Not Enough: Evaluating Confidence Estimation Under Language Variations}

\author{
 \textbf{Yuxi Xia\textsuperscript{1,2}},
 \textbf{Dennis Ulmer\textsuperscript{3}},
 \textbf{Terra Blevins\textsuperscript{4}},
 \textbf{Yihong Liu\textsuperscript{5}},
\\
 \textbf{Hinrich Schütze\textsuperscript{5}},
 \textbf{Benjamin Roth\textsuperscript{1,2}}
\\
 \textsuperscript{1}Faculty of Computer Science, UniVie Doctoral School Computer Science,\\
 \textsuperscript{2}Faculty of Philological and Cultural Studies, University of Vienna, Austria\\
 \textsuperscript{3}ILLC, University of Amsterdam, Netherlands\\
 \textsuperscript{4}Khoury College of Computer Sciences, Northeastern University, USA
 \\
 \textsuperscript{5}LMU Munich, Munich Center for Machine Learning (MCML), Germany
\\
 \small{
   \textbf{Correspondence:} \href{email@domain}{yuxi.xia@univie.ac.at}
 }
}

\begin{document}

\maketitle
\begin{abstract}

Confidence estimation (CE) indicates how reliable the answers of large language models are and impacts user trust and decision-making. 
Existing evaluations mainly concern the alignment between confidence and correctness, but ignore the variability of language: confidence estimates should remain consistent under semantically equivalent prompts or answer variations, while changing when answer meaning differs, as this may indicate a change in correctness.
Therefore, we introduce a novel evaluation framework based on three complementary properties: \textbf{robustness} to prompt perturbations, \textbf{stability} across semantically equivalent answers, and \textbf{sensitivity} to semantically different answers. We show that these metrics are largely independent of existing CE metrics, and that common CE methods often fail on them: while most methods achieve high robustness and stability,  they struggle to distinguish semantically different answers, potentially because they do not effectively leverage generation-side information.
Overall, our framework exposes overlooked limitations of current CE evaluations and provides guidance for selecting confidence estimators for real-world applications.

\end{abstract}

\section{Introduction}

\begin{figure}[tb]
    \centering
    \includegraphics[width=0.985\linewidth]{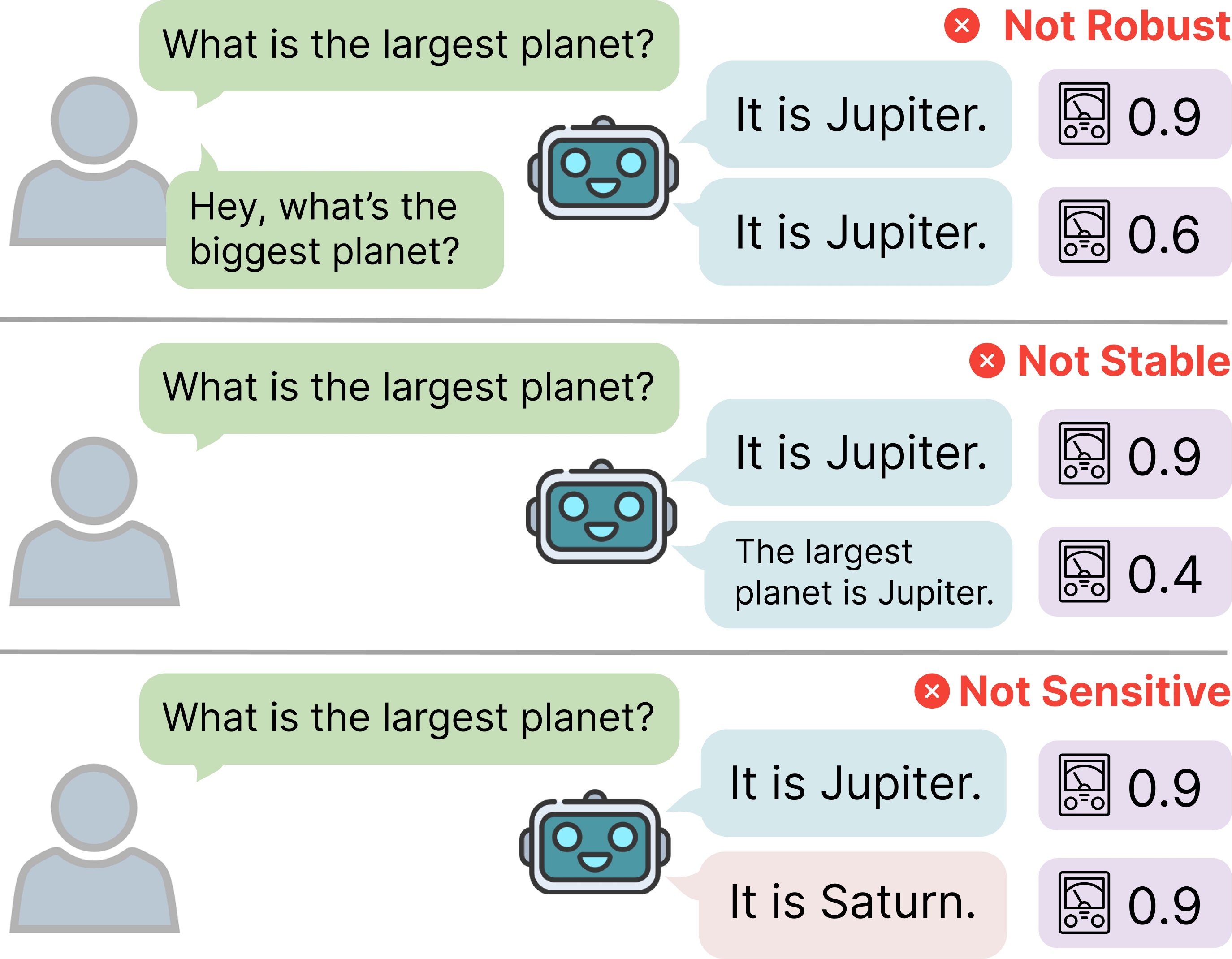}
    \caption{Example of different failure cases of confidence estimators under language variation in either the input prompt or the generated response of the LLM.}
    \label{fig:figure-1}
\end{figure}

Large Language Models (LLMs) are effective at answering fact-based questions across domains such as medicine, law, and education \cite{alqahtani2023emergent, alfertshofer2024sailing, xiao2025llmsassistcomputereducation}. 
In these high-stakes settings, it is paramount to improve user trust and support safer human–AI interactions \cite{kadavath2022language, zhang-etal-2024-calibrating} to avoid harm and achieve positive outcomes.
To this end, \textit{confidence estimation} (CE; \citealp{si2023prompting, tian2023just, chen-mueller-2024-quantifying}) can help by indicating how certain and therefore reliable a model's answer is.

Existing works on CE \cite{calib-guo, osawa2019practical, ovadia2019can, groot-valdenegro-toro-2024-overconfidence, 10.1007/978-981-96-1710-4_10,xiong2024llms} mainly focus on evaluating calibration (whether stated confidences match expected model accuracy), or discrimination (whether it ranks right answers higher than wrong ones).
This overlooks a key property of language (models):
Language is extremely variable, and while the surface form, i.e.\@ a particular choice of words, might change, the meaning of an expression can stay untouched \citep{baan2023uncertainty}. 
However, a method may be well-calibrated yet produce confidence values that fluctuate under minor prompt variations \cite{sclar2023quantifying, xia-etal-2025-influences}, or fail to adapt to answers with different surface forms and / or meanings, as we exemplify in \cref{fig:figure-1}.

This paper presents a novel evaluation paradigm for LLM CE methods that takes language variation into account, including three novel dimensions:
 \textbf{Robustness (P-RB)} measures whether CE methods can ignore semantically irrelevant changes to the prompt and output the same confidence score.
\textbf{Stability (A-STB)} requires  CE methods to assign consistent confidences to semantically equivalent answers, and \textbf{sensitivity (A-SST)} measures whether confidence estimates change accordingly when the answer meaning changes.
In summary, we make four main contributions: (1) we propose three novel and complementary metrics that go beyond calibration to evaluate CE methods under realistic language variation; (2) we conduct a large-scale evaluation of 10 diverse CE methods across 11 LLMs from five model families; (3) we show empirically that while most confidence estimators are robust and stable, they exhibit a striking lack of sensitivity, potentially because generation-side information is not effectively utilized; (4) we provide practical guidance on selecting confidence estimators for target applications and open-source the code.\footnote{Code is available at \url{https://github.com/Yuuxii/Evaluate_confidence_estimation_under_language_variations}.}


\section{Related Work}

\paragraph{Confidence Estimation in LLMs.}
 CE has emerged as a crucial research direction for improving user trust and model interpretability \citep{tian2023just, chen-etal-2024-quantifying}.
Currently, there are 5 common approaches \cite{geng-etal-2024-survey, 10.1145/3744238, shelmanovvashurin2025, testoni-calixto-2026-mind} to estimate LLM confidence: (1) \textit{logit-based}, such as token-level probability \cite{zhu2023calibration, Huang_2025}; (2) \textit{linguistic confidence} \cite{lin2022teaching,xiong2024llms,ulmer2025anthropomimetic, 10.1007/978-3-032-21289-4_34}, which elicits verbalized confidence directly from the model; 
(3) \textit{auxiliary-based} methods \cite{ulmer-etal-2024-calibrating, xia-etal-2025-influences, zhu2026towards}, which employ an auxiliary model trained to predict the confidence of a target LLM based on answer content and predicted correctness; (4) \textit{internal states-based},  \citet{NEURIPS2024_3c1e1fdf} derive model uncertainty from attention heads, other methods \citep{liu2024litcablightweightlanguagemodel, burns2023discovering, kadavath2022language} train probes which comprise smaller, often linear classifiers on internal model representations.
Lastly, (5) \textit{consistency-based methods} \citep{lin2024generating, manakul2023selfcheckgpt, xiong2024llms} rely on multiple stochastic generations obtained by prompting the LLM with loose temperatures and diverse prompts. Confidence is then estimated based on the similarity or frequency of the generated outputs. 
 Since our proposed metrics themselves rely on comparing the confidence of multiple different responses under perturbation, consistency-based methods are not applicable in our framework.

\paragraph{Evaluation of Confidence Estimators.} 
LLMs are typically evaluated using task-level metrics \citep{le-bronnec-etal-2024-exploring, 10.1145/3733567.3735571, xia-etal-2025-learn}, which are not suitable for  CE methods. Instead, calibration metrics, including ECE \citep{calib-guo} and its variants \citep{kumar2019verified, nixon2019measuring, kirchenbauer2022your, blasiok2024smooth}, and Brier score \citep{brier1950verification}, are used to measure the alignment between predicted confidence and actual correctness \citep{groot-valdenegro-toro-2024-overconfidence, zhou-etal-2024-relying, stengeleskin2023calibrated}.
Ranking-based metrics such as AUROC and AUPRC \citep{BRADLEY19971145} further evaluate whether  CE methods can distinguish correct from incorrect responses \citep{vazhentsev2022uncertainty, tian2023just, ulmer-etal-2024-calibrating}, in-distribution from out-of-distribution inputs \citep{ovadia2019can, charpentier2020posterior, ulmer-etal-2022-exploring}, or whether loss correlates with decreased confidence \citep{ulmer-etal-2022-exploring}. However, existing metrics do not assess the robustness of  CE methods to prompt perturbations, as studied for LLM outputs, for instance by ProSA \citep{zhuo-etal-2024-prosa}, or their ability to distinguish between (dis)similar responses.


\section{Background}\label{sec:background}

\paragraph{Notation.} We use $\mathbf{x}$ to denote an input sequence, $\mathbf{t}$ for a language model prompt, and $\mathbf{y}$ for a model response.
We write $\mathbf{y}^*$ for the intended or correct response, and in cases where there are multiple correct responses, we denote the corresponding set by $\mathcal{Y}^*$.
Then, we write a confidence estimator as a function $f$ that predicts a confidence value $c$ conditioned on $\mathbf{x}$, $\mathbf{y}$ and $\mathbf{t}$, written $c = f(\mathbf{y}, \mathbf{x}, \mathbf{t})$.\footnote{
    So e.g.\@ for sequence likelihood, $f$ involves passing $\mathbf{x}$, $\mathbf{t}$ to the LLM to obtain token probabilities. For e.g.\@ an auxiliary model, $f$, can be directly equated with the secondary model.    
}

\subsection{Preliminaries}

One common metric for evaluating CE is the expected calibration error \cite{calib-guo}:
\begin{equation}\label{eq:calibration-error}
    \text{ECE} = \mathbb{E}\big[|p(y = y^* \mid c) - c|\big],
\end{equation}%

\noindent where the expectation is typically approximated by binning $N$ test predictions into $M$ buckets of equal size by their predicted confidence.
The ECE has been criticized for not being a \emph{proper scoring rule}, i.e.\@ trivial predictors other than the true distribution can minimize it \citep{liu2023simple}.
An actual proper scoring rule is the Brier score \cite{brier1950verification}:
\begin{equation}\label{eq:brier-score}
    \text{Brier} = \frac{1}{N}\sum_{i=1}^N\big(c_i - \mathbf{1}(y_i = y^*_i)\big)^2.
\end{equation}

Both ECE and Brier score evaluate whether the confidence score aligns with (expected) correctness. 
However, one might also be interested in whether confidence helps us to distinguish between correct and incorrect predictions. 
This is commonly evaluated through the area under the receiver-operating characteristic curve (AUROC; \citealp{BRADLEY19971145}), where error detection is framed as a binary classification task, and the ratio of the true to false positive rate is measured under all possible thresholds. 
However, none of these metrics incorporates the variability of natural language.

\subsection{An Illustrative Example}\label{sec:example}

To intuitively demonstrate why global calibration metrics fail to catch localized insensitivity to language, we construct a controlled simulation evaluating a baseline that completely ignores answer text.
For each $\mathbf{x}_i$ out of $N$ samples, we obtain $M$ generations $\mathbf{y}_i^{(1)}, \ldots, \mathbf{y}_i^{(M)}$ each.
We sample ``difficulty'' $d_i \sim \mathcal{U}[0, 1]$ for every $\mathbf{x}_i$ and simulate answer correctness $\text{corr}_{i}^{(m)}$ of each $\mathbf{y}_i^{(m)}$ by $\text{corr}_{i}^{(m)} \sim \text{Bernoulli}(1 - d_i)$, such that higher $d_i$ tends to result in a smaller chance of being correct.
We now define confidence estimators, namely an oracle $f_{\text{oracle}}$ which returns the actual correctness: $f_{\text{oracle}}(\mathbf{y}_i^{(m)}) = \text{corr}_{i}^{(m)}$; a constant estimator $f_{\text{constant}}(\mathbf{y}_i) = 1 - d_i$ predicts the difficulty per input, ignoring the generation and its correctness.
Similarly, a prior estimator stoically estimates the expected difficulty across the whole dataset s.t.\@ $f_{\text{prior}}(\mathbf{y}_i) = \mathbb{E}_{d_i \sim \mathcal{U}[0, 1]}[1 - d_i] = 0.5$.
Lastly, a random estimator $f_{\text{random}}(\mathbf{y}_i) \sim \text{Bernoulli}(1 - d_i)$ independently samples a confidence of $1$ or $0$ from the same distribution as $\text{corr}_{i}^{(m)}$.

\begin{table}
    \centering 
    \small
    \renewcommand{\arraystretch}{1.2}
    \resizebox{0.98\linewidth}{!}{
    \begin{tabular}{@{}rccc:cc@{}}
        \toprule
            & ECE $\downarrow$& Brier  $\downarrow$ & AUROC  $\uparrow$ & A-STB $\uparrow$ & A-SST $\uparrow$\\
        \midrule
        $f_{\text{oracle}}$ & 0.00 & 0.00 & 1.00 & 1.00 & 1.00 \\
        $f_{\text{constant}}$ & 0.01 & 0.17 & 0.83 & 1.00 & 0.00 \\
        $f_{\text{random}}$ & 0.33 & 0.33 & 0.67 & 0.72 & 0.18 \\
        $f_{\text{prior}}$ & 0.00 & 0.25 & 0.50 & 1.00 & 0.00 \\
        \bottomrule
    \end{tabular}%
    }
    \caption{Simulation results. Two of our own metrics are stability (A-STB) and sensitivity (A-SST).}\label{tab:simulation}
\end{table}

\paragraph{Results.}
We run a simulation with $N = 100,000$ and $M = 10$ and show the results in \cref{tab:simulation}.\footnote{
For the metrics introduced in \cref{sec:methodology}, we assume that $\mathbf{y}_i^{(m)}$ with the same $\text{corr}_i^{(m)}$ have the same meaning.
}
Unsurprisingly, $f_{\text{oracle}}$ achieves perfect results, and the $f_{\text{random}}$ scores above-chance AUROC, but with the worst calibration.
The $f_{\text{constant}}$ scores perfect ECE, a relatively low Brier score of $0.17$ and an AUROC of $0.83$. 
$f_{\text{prior}}$ shows random-chance AUROC due to only predicting a single value, but reaches perfect ECE and a Brier score of $0.25$, demonstrating that ECE is not a proper scoring rule.
Although these predictors are deliberately simplified, the implications for LLM CE are staggering: 
A random baseline that predicts $0$ and $1$ according to answer difficulty alone can achieve above-random AUROC, but more importantly, $f_{\text{constant}}$, which predicts question difficulty alone and ignores any information from answers, can perform well in both calibration and discrimination. 
A CE method that behaves randomly or that predicts the same confidence, both independently of the actual answer, is uninformative in a practical setting.
There are also two extremes that motivate our novel metrics:
Stability (STB) measures whether an estimator predicts the same confidence for semantically equivalent answers, while sensitivity (SST) captures whether an estimator changes its prediction upon encountering a different answer. 

\begin{figure*}[htpb]
    \centering
    \includegraphics[clip, trim=1cm 0cm 0.4cm 0cm, width=1.\linewidth]{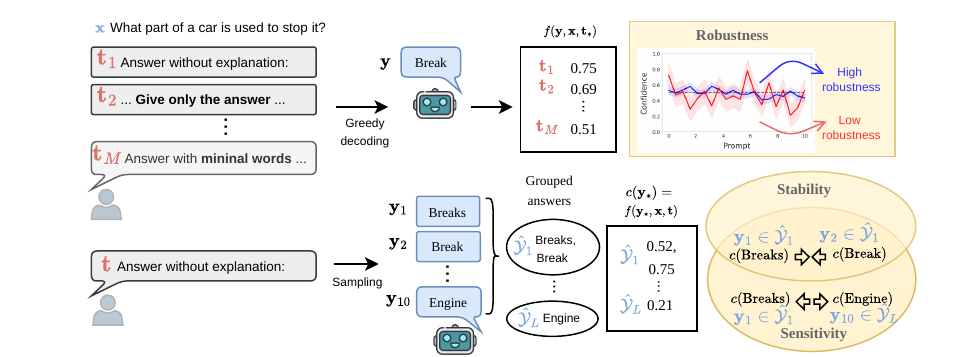}
    \caption{Design of the proposed evaluation metrics:
\textit{Robustness} measures the score variation under semantically irrelevant prompt perturbations. 
\textit{Stability} measures confidence consistency across equivalent answers. \textit{Sensitivity} evaluates whether CE methods assign distinct scores to answers with different meanings than to equivalent ones.}
    \label{fig:prompts}
\end{figure*}
\section{Methodology}\label{sec:methodology}


We investigate whether estimated confidences are reliable based on three qualities:  \textit{robustness} ensures that the estimated confidence is robust towards irrelevant prompt perturbations; \textit{stability} and \textit{sensitivity} quantify how much the CE methods respond to semantically (dis)similar answers.
The computation of all metrics is visualized in \cref{fig:prompts}.
\subsection{Robustness Under Prompt Perturbation}

\paragraph{Motivation.} Prompt formatting is known to influence LLM outputs \citep{min2022rethinking, wang2023label, sclar2023quantifying, feng2024unveiling}, and \citet{xia-etal-2025-influences} shows that it can affect the performance of CE methods. 
In practice, prompts are often modified across applications and users. 
However, while works on verbalized confidence (e.g.\@ \citealp{tian2023just, xiong2024llms}) have varied prompt formats, the robustness of other confidence estimators has not been systematically examined. 
We propose that an ideal CE method should provide consistent confidences when prompts are altered, but responses stay the same:
To isolate prompt variance from generation noise, Robustness is measured by holding the generated response constant while perturbing the instruction format.


\begin{definition}[Robustness]\label[definition]{def:robustness}
Let $\mathcal{T} = \{\mathbf{t}_1,\dots,\mathbf{t}_{M}\}$ be a set of $M$ semantically equivalent prompts.
A confidence estimator $f$ is \emph{robust} to prompt perturbations iff
\[
\forall \mathbf{t}, \mathbf{t}^\prime \in \mathcal{T}, \quad
|f(\mathbf{y}, \mathbf{x}, \mathbf{t}) - f(\mathbf{y}, \mathbf{x}, \mathbf{t}^\prime)| \le \varepsilon.
\]
\end{definition}

\paragraph{Robustness Metric (P-RB$\uparrow$).}
In practice, we measure robustness given a prompt set $\mathcal{T} = \{\textbf{t}_1,\dots,\textbf{t}_M\}$, assuming a ``default'' prompt $\mathbf{t}_1$.
We use a semantic equivalence operation $E(\cdot, \cdot)$ \citep{kuhn2023semantic}\footnote{While \citet{kuhn2023semantic} use a bi-directional entailment classifier to determine semantic equivalence, we use LLM-as-a-judge with the prompt in \cref{sec:prompt}.} to identify the set of all answers $\mathbf{y}_m$ that are equivalent to the answer $\mathbf{y}_1$ using $\mathbf{t}_1$. 
Thus, given $\mathcal{Y} = \{\mathbf{y}^{(m)}\}_{m=1}^M$, we define an index set 
$\hat{\mathcal{I}} = \{m \mid E(\mathbf{y}_1, \mathbf{y}_m) \}$.
Robustness is then computed by measuring the standard deviation $\operatorname{Std}(\cdot)$ of scores across prompts producing equivalent answers and averaging across $N$ samples:
\begin{equation}\label{eq:robustness}
\resizebox{0.99\linewidth}{!}{
$\displaystyle \text{P-RB} = 1 - \frac{1}{N} \sum_{i=1}^{N} \operatorname{Std}\!\left(\left\{f(\mathbf{y}_i^{(m)}, \textbf{x}_i, \textbf{t}_m)\right\}_{m \in \hat{\mathcal{I}}_i}\right),$%
}
\end{equation}

\noindent with $\text{P-RB} > (2 - \varepsilon) / 2$ needed for  \cref{def:robustness} to hold (see \cref{app:theoretical-results}).


\subsection{Stability and Sensitivity Under Answer Variation}

\paragraph{Motivation.}
Setting aside different prompts, LLMs already produce a variety of responses when using stochastic decoding. 
A confidence estimator should incorporate answer semantics:
If the semantics remain unchanged, so should its expected correctness, and thus its confidence.
If the answer differs in meaning, the confidence should be adjusted accordingly.
As we saw in \cref{sec:example}, a constant predictor can be well-calibrated, but is unreliable by ignoring semantic changes.
Conversely, a random predictor is sensitive (since the confidence is random each time), but lacks stability when meaning stays constant.\\

To define these notions more formally, let us again have a set of $M$ generated responses $\mathcal{Y}$ for a given input $\mathbf{x}$ and a (single) prompt $\mathbf{t}$.
Then, given some predicted answer $\mathbf{y}_m \in \mathcal{Y}$, we can define $\hat{\mathcal{Y}} = \{\mathbf{y}^\prime \in \mathcal{Y} \mid E(\mathbf{y}_m, \mathbf{y}^\prime) \}$.
Now we can outline two new desiderata in the face of answer variation:
To evaluate how an estimator handles generation-side variance alone, we freeze the prompt and analyze stochastically sampled outputs.

\begin{definition}[Stability]\label[definition]{def:stability}
    Let $\mathbf{y}^{(m)}$ be one out of $M$ responses given a prompt $\mathbf{t}$, and $\hat{\mathcal{Y}} \subseteq \mathcal{Y}$ a set of semantically equivalent responses to $\mathbf{y}_m$.
    Then, a confidence estimator $f$ is \emph{stable} iff
    \begin{equation}
         \forall \mathbf{y}^\prime \in \hat{\mathcal{Y}}:\quad |f(\mathbf{y}^{(m)}, \mathbf{x}, \mathbf{t}) - f(\mathbf{y}^\prime, \mathbf{x}, \mathbf{t})| \le \varepsilon.\nonumber
    \end{equation} 
\end{definition}

From this, we define the complementary notion of sensitivity, namely that semantically different answers should receive different scores: 

\begin{definition}[Sensitivity]\label[definition]{def:sensitivity}
    Let $\mathcal{E}(\mathcal{Y}) = \{\hat{\mathcal{Y}}_l\}_{l=1}^L$ be a finite partition of $\mathcal{Y}$ s.t.\@ $\bigcup_{l=1}^L \hat{\mathcal{Y}}_l = \mathcal{Y}$ and $\forall \hat{\mathcal{Y}}_l, \hat{\mathcal{Y}}_{l^\prime} \in \mathcal{E}, l\neq l^\prime: \hat{\mathcal{Y}}_l \cap \hat{\mathcal{Y}}_{l^\prime} = \varnothing$.
    Then, given an input $\mathbf{x}$, prompt $\mathbf{t}$ a set of semantically equivalent responses $\hat{\mathcal{Y}}_l \in \mathcal{E}(\mathcal{Y})$, a confidence estimator $f$ is \emph{sensitive} iff 
    \begin{align}
    & \forall \hat{\mathcal{Y}}_l, \hat{\mathcal{Y}}_{l^\prime} \in \mathcal{E}(\mathcal{Y}), l\neq l^\prime, \mathbf{y} \in \hat{\mathcal{Y}}_l, \mathbf{y}^\prime \in \hat{\mathcal{Y}}_{l^\prime}:\nonumber\\
    & |f(\mathbf{y}, \mathbf{x}, \mathbf{t}) - f(\mathbf{y}^\prime, \mathbf{x}, \mathbf{t})| > \varepsilon.\nonumber
    \end{align}

\end{definition}
 
\paragraph{Stability Metric (A-STB$\uparrow$).}
Practically, we do not check all sets $\hat{\mathcal{Y}}_l$, but just restrict ourselves to the most meaningful ones:
 For each sample, the largest set of answers is denoted as $\hat{\mathcal{Y}}^\text{max}$, while the set with the fewest ones is $\hat{\mathcal{Y}}^\text{min}$.
We then again compute the standard deviation over confidence scores among semantically equivalent answers in $\hat{\mathcal{Y}}^\text{max}$ over $N$ data points:
\begin{equation}\label{eq:stability}
\resizebox{0.99\linewidth}{!}{
$\displaystyle
\text{A-STB}
= 1 -
\frac{1}{N}
\sum_{i=1}^{N}
\operatorname{Std}\!\left(
\left\{
f(\mathbf{y}_j, \mathbf{x}_i, \mathbf{t})
\right\}_{\mathbf{y}_j  \in \hat{\mathcal{Y}}^\text{max}_i}
\right).$%
}
\end{equation}
Applying the same argument as for robustness, we show in \cref{app:theoretical-results} that $\text{A-STB} > (2 - \varepsilon)/2$ must hold for \cref{def:stability}.

\begin{figure}[tb]
    \centering
    \includegraphics[clip, trim=0cm 0cm 0.4cm 0.5cm, width=.95\linewidth]{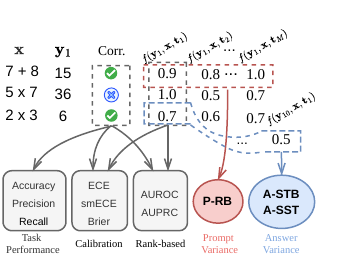}
    \caption{Comparison of the information CE evaluation metrics require. While others need correctness labels, P-RB, A-STB and A-SST solely rely on multiple answers under different prompts or stochastic decoding.}
    \label{fig:compare}
\end{figure}

\paragraph{Sensitivity Metric (A-SST$\uparrow$)}
To quantify how effectively a confidence estimator distinguishes semantically different answers relative to semantically equivalent ones, we compute
\begin{equation}\label{eq:sensitivity}
\resizebox{0.99\linewidth}{!}{
      $\displaystyle\text{A-SST} = \frac{1}{N}\sum_{i=1}^N |\Delta (\hat{\mathcal{Y}^\text{max}_i}, \hat{\mathcal{Y}}^\text{min}_i) -\Delta (\hat{\mathcal{Y}^\text{max}_i}, \hat{\mathcal{Y}}^\text{max}_i)|,$%
      }
\end{equation}
\noindent where, using $c(\mathbf{y}) \equiv f(\mathbf{y}, \mathbf{x}, \mathbf{t})$ as a shorthand,
\begin{align}
    \Delta (\hat{\mathcal{Y}}, \hat{\mathcal{Y}}^\prime) & = \frac{1}{|\hat{\mathcal{Y}}||\hat{\mathcal{Y}}^\prime|}\sum_{\mathbf{y} \in \hat{\mathcal{Y}}}\sum_{\mathbf{y}^\prime \in \hat{\mathcal{Y}}^\prime}\! |c(\mathbf{y})-c(\mathbf{y}^\prime)| .
\end{align}
When the confidence estimator is both stable and sensitive according to \cref{def:stability,def:sensitivity}, then this metric is bounded by $\text{A-SST} > \varepsilon / N \sum_{i=1}^N |\hat{\mathcal{Y}}^\text{max}_i|^{-1} $, as we derive in \cref{app:theoretical-results}.
A high value indicates that the score differences between semantically different answers are larger than among equivalent answers, and the confidence estimator is therefore sensitive.
It should be noted that our metrics do not require any information about the correctness of a response, in contrast to other common metrics in \cref{fig:compare}.
This is especially attractive in cases where automated correctness judgments are expensive, difficult or noisy, e.g.\@ ambiguous questions \citep{min2020ambigqa} or time-sensitive questions \citep{pletenev-etal-2025-will}.

\section{Experiments}\label{sec:experiments}

In our experiments section, we aim to answer the following three research questions:

\paragraph{\textbf{RQ1}:}
How do our proposed metrics relate to established metrics for the evaluation of confidence estimators? Do they measure something different?

\paragraph{\textbf{RQ2}:}
Prior research has shown that calibration in LLMs improves with scale \citep{minderer2021revisiting, dan2021effects, kadavath2022language, zhu2023calibration, xiong2024llms, tao2025revisiting}. Is this also true for our metrics?

\paragraph{\textbf{RQ3}:}
Combining all metrics, which confidence estimators perform best across different model families and datasets? 

\subsection{Setup}

Our experiments are conducted on an aggregation of four question-answering datasets and 11 LLMs across different model families and sizes.


\paragraph{Evaluated Methods.}  We benchmark 10 CE methods across 4 categories: (1) Output logit-based: Sequence likelihood (\emph{Seq.\@ Likelihood}) aggregates the token probabilities of the generated answer. \emph{Platt Scaling} (\citealp{platt1999probabilistic}) learns a scale and shift scalar to calibrate sequence likelihood (in our case on the training set). \emph{Boosted Prob.\@} by \citet{dinh-niehues-2025-generative} identifies dominant token probabilities in the output sequence. \emph{P(True)} prompts the LLMs to judge if the response is true, and uses the normalized probability of the ``True'' token \cite{kadavath2022language}. (2) Linguistic features: \emph{Verbalized Conf.} \citep{tian2023just, xiong2024llms} prompts the LLM to state how certain it is about its answer.  (3) Auxiliary models: Calib-1-Focal (\emph{Calib1}) trains an auxiliary classifier \citep{ulmer-etal-2024-calibrating, xia-etal-2025-influences}, 
uses response correctness as training targets. (4) Internal states-based: \emph{Attention Score} and \emph{Hidden Score} \citep{NEURIPS2024_3c1e1fdf} are derived from the mean log-determinant of attention heads and hidden representations, respectively;  \emph{P(IK)} \citep{kadavath2022language} and \emph{SAPLMA} \citep{azaria-mitchell-2023-internal} train probe classifiers on last-layer hidden states from different token positions. Implementation details are in \cref{implement-detail}. All the confidence scores estimated by different methods are scaled to 0-1 for evaluation.

\paragraph{Datasets.} We test on four open-ended question-answering benchmarks covering diverse topics. 
They include 2,000 randomly sampled training examples each from Natural Questions (NQ; \citealp{kwiatkowski2019natural}), SciQ \cite{welbl-etal-2017-crowdsourcing}, TriviaQA \cite{joshi2017triviaqa}, and PopQA \cite{mallen-etal-2023-trust}. 
All methods are evaluated on 1,000 randomly sampled test questions each. 

\paragraph{LLMs.}  We evaluate 11 LLMs spanning five major families, including Mistral \citep{mistral2024largeinstruct, mistralai2025ministral3reasoning8b}, Llama 3 \citep{meta2024llama3}, Qwen 2.5 \citep{qwen2024qwen2.5}, OLMo 2 \citep{olmo2024furious} and GPT-4o \citep{openai2024gpt4o} models, whose known sizes range from 7B to 123B and include both instruction and reasoning models. 
More details are given in \cref{app:llm-detail}. 

\paragraph{Experimental Design.} 
For measuring prompt robustness, we use greedy decoding to ensure that any variability is solely driven by prompt perturbations, for which we test the $10$ variants shown in \cref{tab:ans_prompt_detail} (\cref{sec:ans-prompt}). 
To avoid introducing ambiguity, we keep the question fixed and perturb only the instructions. 
Variants include forms such as ``answer the question'' and ``provide an answer to the question.'' 
Each prompt also enforces conciseness, with phrases like ``give ONLY the answer,'' ``no other words or explanation,'' etc.
For stability and sensitivity, we set the model temperature to 0.7 to generate $10$ sampled answers.

 We use GPT-4o as a judge model (see the judge prompt in \cref{prompt:correctness} and \cref{prompt:grouping} in the appendix) to determine the semantic equivalence between LLM responses and target answers for correctness evaluation, as well as the semantic equivalence among sampled responses when computing stability and sensitivity metrics.

\subsection{Comparing Evaluation Metrics}

\begin{figure}[tb]
    \centering
    \begin{subfigure}[t]{0.95\linewidth}
        \centering
        \includegraphics[width=0.985\linewidth]{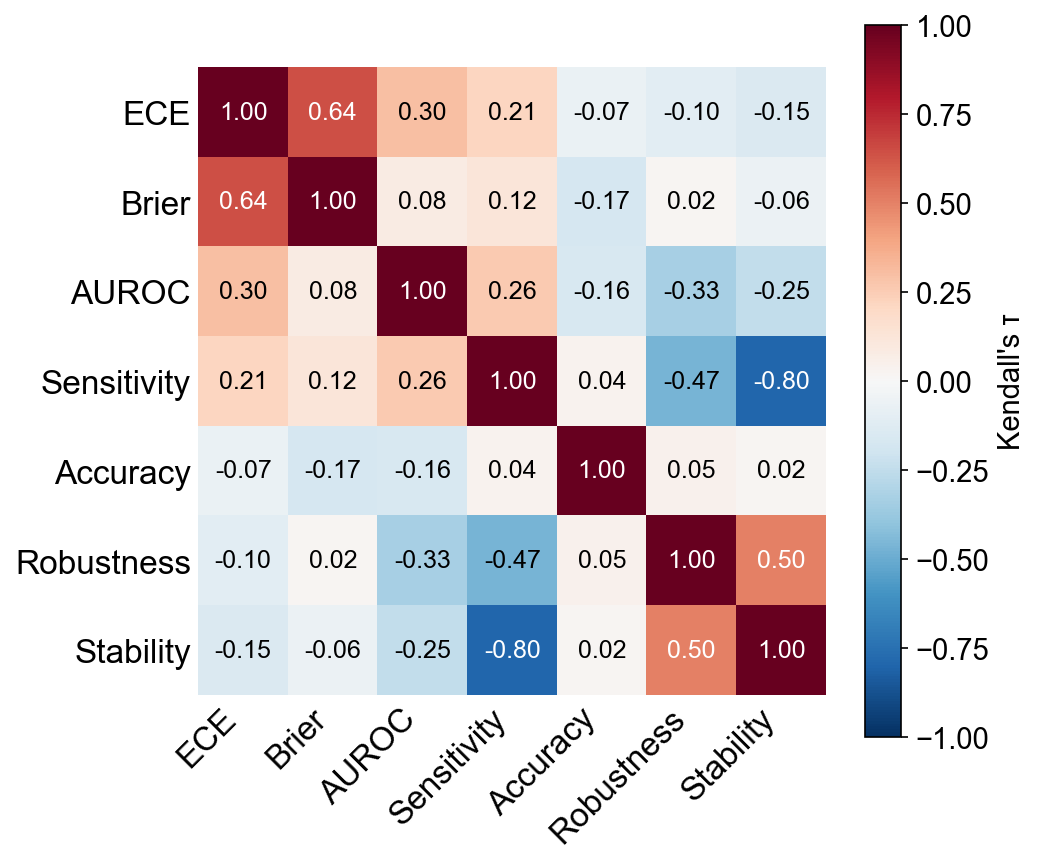}
        \caption{Kendall's $\tau$ between metrics. Statistic test results are shown  in \cref{sec:significance}.}\label{subfig:kendalls-tau}
    \end{subfigure}%
    
    \begin{subfigure}[t]{0.95\linewidth}
        \centering
        \includegraphics[width=0.95\linewidth]{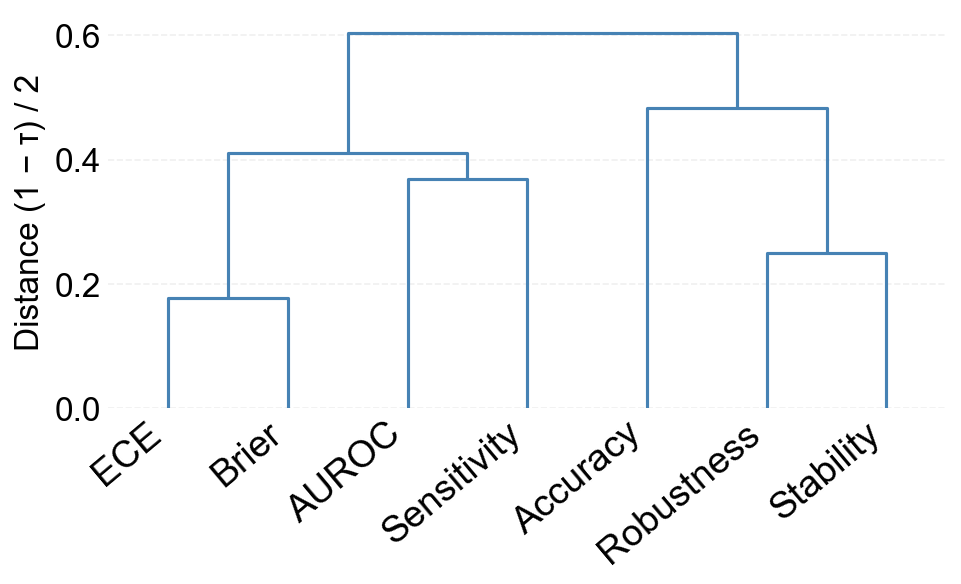}
        \caption{Dendrogram of hierarchical clusters.}\label{subfig:dendrogram}
    \end{subfigure}
    \caption{Similarity of CE evaluation metrics. (a) Heatmap of Kendall's $\tau$ between metrics, including accuracy. (b) Hierarchical clustering based on $\tau$ values.}\label{fig:metric-comparison}
\end{figure}

We first compare the similarity of CE evaluation metrics by pooling results across models and confidence estimators, computing Kendall's $\tau$ \citep{kendall1938new} correlation coefficient between them, and then averaging the result across our four datasets.
We then run hierarchical clustering over the results, treating correlation (as $(1 - \tau)/2$) as a distance measure to identify closely related metrics. 

\begin{figure*}[thb!]
    \centering
    \includegraphics[width=\textwidth]{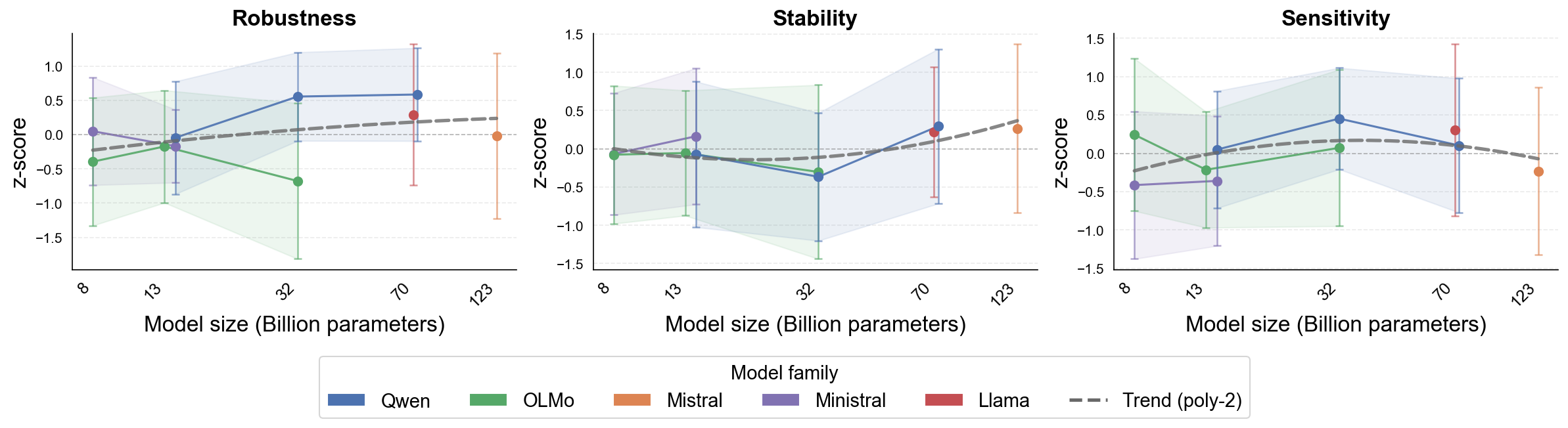}
    \caption{Normalized metric value distributions by model family and size. 
    Whiskers and shaded regions indicate the standard deviation over observed metric values across datasets and CE methods. 
    Models of the same family are connected and appear in the same color. Trend lines are second-degree polynomials fitted on all observations.}\label{fig:scaling-laws-shaded-ours}
\end{figure*}

\begin{table*}[htbp]
\centering\footnotesize
\sisetup{round-mode=places, round-precision=2, table-format=1.2, table-number-alignment=center,detect-weight=true,
  detect-family=true}
\setlength{\tabcolsep}{2pt}
\renewcommand{\arraystretch}{1.1}
\resizebox{\linewidth}{!}{%
\begin{tabular}{@{}c l *{6}{S} @{\hspace{1.2em}} *{6}{S}@{}}
\toprule
& & \multicolumn{6}{c}{\textbf{SciQ}} & \multicolumn{6}{c}{\textbf{PopQA}}\\
\cmidrule(lr){3-8} \cmidrule(lr){9-14}
& Method
& {ECE$\downarrow$} & {Brier$\downarrow$} & {AUROC$\uparrow$} & {P-RB$\uparrow$} & {A-STB$\uparrow$} & {A-SST$\uparrow$}
& {ECE$\downarrow$} & {Brier$\downarrow$} & {AUROC$\uparrow$} & {P-RB$\uparrow$} & {A-STB$\uparrow$} & {A-SST$\uparrow$} \\
\midrule
\multirow{4}{*}{\rotatebox[origin=c]{90}{\scriptsize Logit-based}}
  & Seq. Likelihood & 0.077 & \bfseries 0.150 & 0.683 & 0.938 & 0.919 & 0.167 & 0.386 & 0.309 & 0.841 & 0.951 & 0.936 & 0.196 \\
  & Boosted Prob.   & 0.176 & 0.180 & 0.725 & \bfseries 0.994 & \bfseries 0.996 & 0.011 & 0.575 & 0.510 & 0.862 & 0.979 & 0.973 & 0.072 \\
  & Platt Scaling   & 0.096 & 0.156 & 0.683 & \bfseries 0.987 & 0.981 & 0.039 & 0.277 & 0.253 & 0.841 & \bfseries 0.990 & 0.987 & 0.040 \\
  & P(True)         & 0.153 & 0.167 & 0.729 & 0.953 & 0.970 & 0.183 & 0.329 & 0.316 & 0.797 & 0.928 & 0.938 & \bfseries 0.251 \\
\cmidrule(l){2-14}
\multirow{4}{*}{\rotatebox[origin=c]{90}{\scriptsize Internal States}}
  & Attention Score & \bfseries 0.017 & \bfseries 0.152 & 0.552 & \bfseries 0.991 & 0.993 & 0.019 & \bfseries 0.051 & 0.199 & 0.614 & \bfseries 0.987 & 0.981 & 0.046 \\
  & Hidden Score    & 0.037 & \bfseries 0.150 & 0.588 & 0.980 & 0.982 & 0.044 & \bfseries 0.047 & 0.203 & 0.569 & \bfseries 0.993 & 0.986 & 0.032 \\
  & SAPLMA          &  0.154 & 0.170 & \bfseries 0.745 & 0.893 & 0.923 & \bfseries 0.209 & 0.137 & \bfseries 0.152 & \bfseries 0.876 & 0.923 & 0.947 & 0.103 \\
  & P(IK)           & 0.062 & \bfseries 0.148 & 0.684 & 0.976 & \bfseries 1.000 & 0.000 & 0.080 &  0.155 & 0.824 & 0.967 & \bfseries 1.000 & 0.000 \\
\cmidrule(l){2-14}
\multirow{1}{*}{\rotatebox[origin=c]{90}{\hspace{1cm}\scriptsize Ling.\hspace{-0.1cm}}}
  & Verbalized Conf.& 0.147 & 0.168 & 0.636 & \bfseries 0.990 & \bfseries 0.995 & 0.033 & 0.610 & 0.585 & 0.657 & 0.984 & 0.978 & 0.072 \\
\cmidrule(l){2-14}
\multirow{1}{*}{\rotatebox[origin=l]{90}{\scriptsize\shortstack{Aux.}\hspace{-0.1cm}}}
  & Calib1          & 0.095 & 0.160 & 0.601 & \bfseries 0.985 & 0.971 & 0.067 & 0.087 & 0.167 & 0.778 & \bfseries 0.985 & 0.974 & 0.073 \\
\bottomrule
\end{tabular}%
}
\caption{Calibration and our metrics averaged across all models. Arrows indicate metric direction ($\downarrow$ lower is better, $\uparrow$ higher is better); the best value per column is bold. Results of other datasets are shown in \cref{tab:nq-triviaqa} in \cref{sec:nq-triviaqa}.} \label{tab:sciq-popqa}
\end{table*}

\paragraph{Results.} 
\cref{subfig:kendalls-tau} shows only a mild correlation between our and established metrics:
We find a $\tau = 0.26$ for sensitivity and AUROC, which is much lower than e.g.\@ $\tau = 0.64$ between Brier score and ECE.
In contrast, our metrics' correlation coefficients with each other range from $-0.80$ to $0.50$.
Interestingly, no metric has a meaningful correlation with accuracy.
Using the hierarchical clustering in \cref{subfig:dendrogram}, we can further observe that robustness and stability are clustered together, albeit their similarity is still trumped by ECE and Brier score; sensitivity remains grouped with AUROC.
In \cref{fig:boxplots-datasets} in \cref{sec:distribution-add}, we plot metric distributions, showing strongly skewed results for robustness and stability on all datasets, contrasted by an overall lack of sensitivity; all the while, distributions appear very distinct from ECE, AUROC, and Brier score.
Addressing \textbf{RQ1}, we take this as evidence that our metrics measure novel and so-far overlooked properties of confidence estimators.

\subsection{Scaling Laws for Confidence Estimates}

Prior literature has repeatedly shown that larger models tend to be better calibrated \citep{ahuja2022calibration, zhu2023calibration, xiong2024llms, tao2025revisiting}.
To test whether the same is true for our metrics, we plot the distribution of metric scores per model across datasets and CE methods, using parameter size in billions as a reference point.\footnote{We remove Calib1 since it uses an auxiliary model.}
Since metric score distributions differ between confidence estimators and datasets, we apply $z$-score normalization across measurements for the same CE method and dataset from the same model family.
This isolates the general improvement of metric scores, even if dataset difficulty and the performance of specific CE methods differ.

\paragraph{Results.}
We show trends for our metrics in \cref{fig:scaling-laws-shaded-ours}, demonstrating a mild improvement with model size.\footnote{
    \cref{fig:scaling-laws-shaded-calib} (\cref{sec:scaling-add}) shows the plots for ECE, Brier score, and AUROC, where we mostly corroborate existing findings.
    The weaker improvements with scale do not contradict prior work, since often only specific estimators such as token/sequence likelihood \citep{zhu2023calibration, tao2025revisiting} or verb.\@ confidence \citep{kadavath2022language} are considered.
}
This trend by estimator is often inconsistent depending on the type of method and model, as we show in more detailed graphs in \cref{fig:scaling-laws-method-families} in \cref{sec:scaling-add}.
For \textbf{RQ2}, scaling may improve confidence quality with respect to language variation; however, the substantial differences across estimators and models highlight the need for careful evaluation in practical settings.

\begin{figure}[htb]
    \centering
    \includegraphics[width=0.95\linewidth]{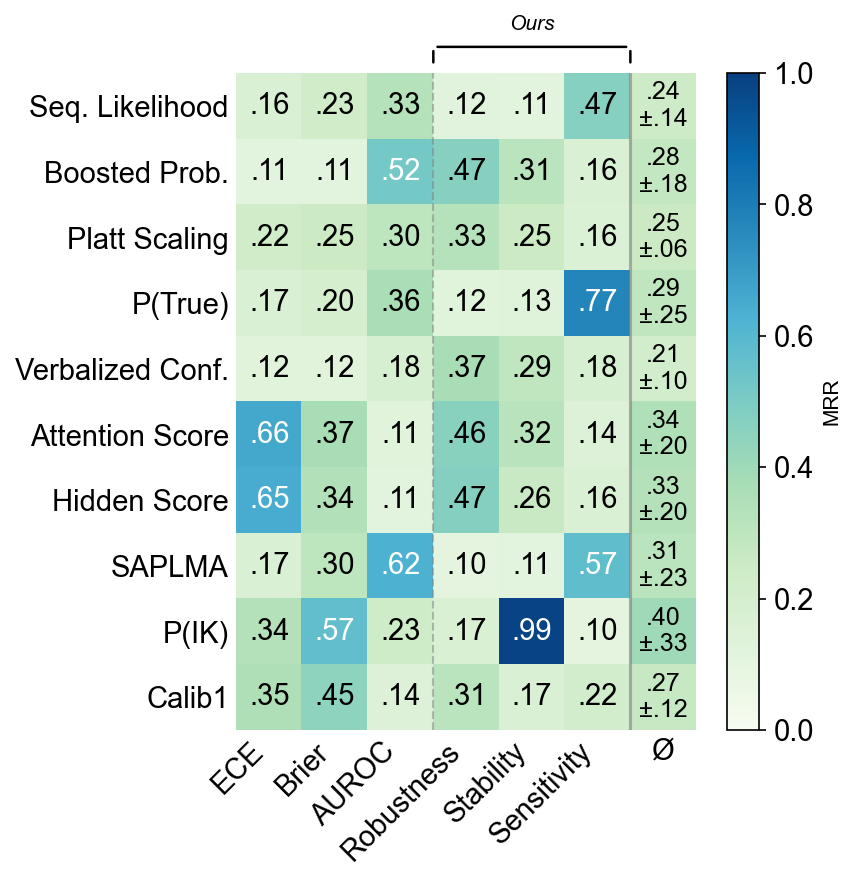}
    \caption{Mean reciprocal rank of CE methods per evaluation metric, averaged over datasets. $\varnothing$ is another average across all metrics, including standard deviation.}
    \label{fig:mrr-summary}
\end{figure}
\subsection{Comparing Confidence Estimators}\label{sec:comparing-ce}

To test whether any confidence estimators reign supreme across all models and datasets, we first analyze the evaluation results per metric and dataset, averaged across models.
For each CE method, we then compute the mean reciprocal rank (MRR), where $1$ means that it outperformed all others.

\paragraph{Results.} 
\cref{tab:sciq-popqa} shows the results for SciQ and PopQA, with the remaining dataset results shown in \cref{tab:nq-triviaqa} in \cref{sec:nq-triviaqa}.
Overall, no method outperforms any other on all metrics. 
Well-calibrated methods also seem to struggle with sensitivity and vice versa.
At the same time, while all confidence estimators score high on robustness (P-RB) and stability (A-STB), most fail on sensitivity (A-SST).
By overall MRR, as shown in \cref{fig:mrr-summary} across all datasets and in \cref{fig:mrr-per-dataset} in \cref{sec:rank-add} per dataset, $P\text{(IK)}$ achieves the best overall MRR, but with low ranking on robustness and sensitivity. 
Regarding \textbf{RQ3}, this highlights that the choice of the best CE method will be highly dependent on the specific dataset, model, and other constraints, such as different priorities of certain metrics.

\subsection{Why Is Sensitivity So Low?}\label{sec:why-sens}

In the previous section, we found that almost all CE methods scored high on robustness and stability, but very low on sensitivity. 
This means that in practice, confidence estimates fail to adapt to semantically \emph{different} responses. 
This is worrying, since a different response likely also implies a change in output quality or correctness, and therefore a miscalibrated confidence. 
To investigate this phenomenon, we display specific responses in detail (see \cref{tab:case-stab-sens-detail,tab:case-stab-sens-detail-3,tab:case-stab-sens-detail-19} in \cref{sec:case-study}).
We hypothesize that CE methods leveraging information closer to the response generation exhibit higher sensitivity to answer changes. 
Estimators trained solely on the question's final token (like $P\text{(IK)}$) are inherently blind to the generated text, resulting in low sensitivity; in contrast, methods evaluating the response's final token (like SAPLMA) achieve much higher sensitivity. 
Similarly, methods that directly rely on output logits without post-hoc scaling, such as sequence likelihood and P(True), tend to be more sensitive to changes in the generated answer, as shown by the rankings in \cref{fig:mrr-summary}.
This aligns with results by \citet{eusebi2026reading}, showing that correctness information is incrementally encoded during generations.

%

\section{Discussion}
Since our results in \cref{sec:comparing-ce} found no universally superior confidence estimator, the choice of method depends on the target application, as different settings require trade-offs between robustness, stability, and sensitivity. 
For instance, user-facing QA systems prioritize robustness to maintain consistent abstention behavior under evolving prompts. 
In contrast, multi-candidate reranking settings, such as LLM-as-a-judge \citep{wagner2024black, jung2025trust} or reward models \citep{yang2024bayesian}, rely more on stability and sensitivity to assess competing responses. 
Conditional computation and model cascades \citep{gupta2024language} emphasize sensitivity to detect low-confidence predictions for escalation to stronger models. 
Multi-agent collaboration \citep{chen2024reconcile, yang2024confidence} requires balancing all three properties, while in controlled environments with fixed prompts and limited answer variation, traditional calibration metrics are sufficient.
Importantly, however, the observed lack of sensitivity across all confidence estimators implies that methods should better exploit relevant information across generation time steps \citep{eusebi2026reading} and model components \citep{shelmanov2025head}.

\section{Conclusion}

We introduced three novel complementary metrics (robustness, stability and sensitivity) to evaluate whether confidence estimates remain reliable under language variation. 
Our results show that these metrics are largely independent of traditional CE metrics for calibration, suggesting that existing evaluations capture only a limited view of confidence quality.
While most CE methods achieve high calibration, robustness and stability, they remain insensitive to changes in answer meaning.
This raises concerns for downstream applications such as multi-candidate reranking.
We further show that scaling alone does not resolve this issue, and our findings suggest that incorporating more answer-side information may improve sensitivity.

As no CE method consistently dominates, we highlight fundamental trade-offs between these metrics, suggesting that the ``best'' confidence estimator depends strongly on the target application.

\section*{Limitations}
While our study provides a comprehensive evaluation of LLM confidence estimation across robustness, stability, sensitivity, calibration, and discrimination, several limitations remain.

First, our analysis is limited to English and fact-based question answering tasks; extending to multilingual or open-ended domains (e.g., reasoning, dialogue, multiple-choice, or creative generation) may reveal different findings.
Second, although we designed a diverse set of prompting strategies, they do not exhaustively capture all linguistic and contextual variations that may influence confidence expression. In particular, it remains challenging to construct more aggressive prompt perturbations that preserve identical semantic meaning and avoid ambiguity across all questions and datasets.
Third, part of our evaluation relies on GPT-4o as a judge model to assess answer correctness and semantic equivalence. While this enables scalable annotation, it may introduce model-specific biases \citep{bavaresco2025llms}.
Fourth, we do not include consistency-based confidence estimation methods, since our metrics themselves rely on multiple samples to be computed, and are therefore incompatible with self-consistency. Consistency-based methods are not naturally superior on our metrics; for instance, they could easily assign the same confidence score to two semantically different answers appearing at the same frequency in the sampling pool.  
In the end, our contribution is not comprehensively benchmark all existing CE methods, but instead to highlight overlooked aspects regarding language variation using our proposed metrics.

\section*{Acknowledgments} 
This research has been funded by the Vienna Science and Technology Fund (WWTF)[10.47379/VRG19008] ``Knowledge infused Deep Learning for Natural Language Processing'', and is supported by the Dutch National Science Foundation (NWO Vici VI.C.212.053). We acknowledge the support of the OpenAI Researcher Access Program for providing API credits used in this research.

\input{}

\bibliography{custom}

\newpage

\appendix
\begin{CJK}{UTF8}{gbsn}
\section{Theoretical Results}\label[appendix]{app:theoretical-results}

In this section, we directly relate the definitions we lay out in \cref{sec:methodology} to our proposed metrics. 
To this end, we formally prove lower bounds that our metrics have to exceed in order for the definitions to hold.
We start with \cref{def:robustness} and the prompt robustness metric in \cref{eq:robustness}.

\begin{theorem}\label{theorem:robustness}
    Given a set of $M$ semantically equivalent prompts  $\mathcal{T} = \{\mathbf{t}_1,\dots,\mathbf{t}_{M}\}$ and a \emph{robust} confidence estimator $f$ according to \cref{def:robustness} s.t.\@
    \begin{equation}\label{eq:repeated-def}
    \forall \mathbf{t}, \mathbf{t}^\prime \in \mathcal{T}, \quad
    |f(\mathbf{y}, \mathbf{x}, \mathbf{t}) - f(\mathbf{y}, \mathbf{x}, \mathbf{t}^\prime)| \le \varepsilon,
    \end{equation}
    \noindent then it must hold for the robustness metric $\hat{\sigma}_r$ defined in \cref{eq:robustness} that
    \begin{equation}
        \text{P-RB} > \frac{2 - \varepsilon}{2}.
    \end{equation}
    
\end{theorem}

\begin{proof}
    In the following, we will use the shorthand $c$ for $f(\mathbf{y}, \mathbf{x}, \mathbf{t})$.
    Because of \cref{def:robustness}, we know that any two $c, c^\prime$ cannot lie more than $\varepsilon$ apart.
    Since standard deviation and variance are metrics of dispersion, they are invariant to translation, meaning that they will stay constant if the data is shifted by a constant amount.
    Therefore, let us assume that $c \in [0, \varepsilon]$.
    Since we would like to find the maximum dispersion possible under \cref{eq:repeated-def}, let us further assume that $k$ points are equal to $0$, and $M^\prime - k$ equal $\varepsilon$,
    where $M^\prime \le M$ refers to number of generations that are equivalent to the default generation $\mathbf{y}_1$, including $\mathbf{y}_1$.
    Therefore, their mean becomes $\mu = \frac{M^\prime-k}{M^\prime}\varepsilon$.
    Correspondingly, we can evaluate their variance as 
    \begin{align}
        \sigma^2 & = \frac{1}{M^\prime}[k(0-\mu)^2 + (M^\prime - k)(\varepsilon - 
    \mu)^2] \\
        & = \frac{1}{M^\prime}\Big[ k\frac{(M^\prime-k)^2\varepsilon^2}{{M^\prime}^2} + (M^\prime-k)\frac{k^2\varepsilon^2}{{M^\prime}^2} \Big] \\
        & = \frac{\varepsilon^2}{{M^\prime}^3}[k(M^\prime -k)^2 + (M^\prime - k)k^2] \\
        & = \frac{\varepsilon^2}{{M^\prime}^3}k(M^\prime-k)M^\prime.
    \end{align}
    Next, we would like to identify the $k$ that maximizes $k(M^\prime-k)M^\prime$, we can easily be identified as $k = \frac{M^\prime}{2}$. 
    Reinserting it into the previous expression yields
    \begin{align}
        & = \frac{\varepsilon^2}{{M^\prime}^3}\frac{{M^\prime}}{2}\Big({M^\prime}-\frac{M^\prime}{2}\Big)M^\prime \\
        & = \frac{\varepsilon^2}{{M^\prime}^3}\frac{{M^\prime}^3}{4} = \frac{\varepsilon^2}{4}
        \leftrightarrow \sigma = \frac{\varepsilon}{2}.
    \end{align}
    By plugging this result into the definition of robustness in \cref{def:robustness}, we can show that since the biggest standard deviation for which \cref{eq:repeated-def} still holds is $\frac{\varepsilon}{2}$ leads robustness to amount to 
    \begin{equation}
        \text{PR-B} > 1 - \frac{\varepsilon}{2} = \frac{2 - \varepsilon}{2}.
    \end{equation}
    
\end{proof}

Since \cref{def:stability} and the stability metric in \cref{eq:stability} follow the same structure, the matching theorem follows naturally.

\begin{theorem}
    Given a set of $M$ responses given a prompt $\mathbf{t}$, $\hat{\mathcal{Y}} \subseteq \mathcal{Y}$ a set of semantically equivalent responses to $\mathbf{y}_m$ and a \emph{stable*} confidence estimator $f$ according to \cref{def:stability} s.t.\@
    \begin{equation}
        \forall \mathbf{y}^\prime \in \hat{\mathcal{Y}}:\quad |f(\mathbf{y}^{(m)}, \mathbf{x}, \mathbf{t}) - f(\mathbf{y}^\prime, \mathbf{x}, \mathbf{t})| \le \varepsilon.\nonumber
    \end{equation}
    Then it must hold for the stability metric $\hat{\sigma}_s$ defined in \cref{eq:robustness} that
    \begin{equation}
        \text{A-STB} > \frac{2 - \varepsilon}{2}.
    \end{equation}
\end{theorem}

\begin{proof}
The proof follows the same argument as \cref{theorem:robustness} without loss of generality.
\end{proof}

Lastly, we would like to prove a similar bound for the sensitivity metric in \cref{eq:sensitivity}, which we do below. 

\begin{theorem}
    Let $\mathcal{Y}$ denote a set of $M$ generated responses for a given input $\mathbf{x}$ and a (single) prompt $\mathbf{t}$ s.t.\@ $\mathcal{Y} = \{\mathbf{y}^{(1)}, \ldots, \mathbf{y}^{(M)}\}$. 
    Also let $\mathcal{E}(\mathcal{Y}) = \{\hat{\mathcal{Y}}_l\}_{l=1}^L$ be a finite partition of $\mathcal{Y}$ s.t.\@ $\bigcup_{l=1}^L \hat{\mathcal{Y}}_l = \mathcal{Y}$ and $\forall \hat{\mathcal{Y}}_l, \hat{\mathcal{Y}}_{l^\prime} \in \mathcal{E}, l\neq l^\prime: \hat{\mathcal{Y}}_l \cap \hat{\mathcal{Y}}_{l^\prime} = \varnothing$ and let $f$ be a \emph{sensitive} confidence estimator according to \cref{def:sensitivity} s.t.\@
    \begin{align}
        & \forall \hat{\mathcal{Y}}_l, \hat{\mathcal{Y}}_{l^\prime} \in \mathcal{E}(\mathcal{Y}), l\neq l^\prime, \mathbf{y} \in \hat{\mathcal{Y}}_l, \mathbf{y}^\prime \in \hat{\mathcal{Y}}_{l^\prime}:\nonumber\\
        & |f(\mathbf{y}, \mathbf{x}, \mathbf{t}) - f(\mathbf{y}^\prime, \mathbf{x}, \mathbf{t})| > \varepsilon, 
    \end{align}
    \noindent and that $f$ is stable according to \cref{def:stability} s.t.\@
    \begin{equation}
        \forall \mathbf{y}^\prime \in \hat{\mathcal{Y}}:\quad |f(\mathbf{y}^{(m)}, \mathbf{x}, \mathbf{t}) - f(\mathbf{y}^\prime, \mathbf{x}, \mathbf{t})| \le \varepsilon.\nonumber
    \end{equation}
    Then it must hold for the sensitivity metric A-SST defined in \cref{eq:sensitivity} that
    \begin{equation}
        \text{A-SST} > \frac{\varepsilon}{N} \sum_{i=1}^N |\hat{\mathcal{Y}}^\text{max}_i|^{-1}.
    \end{equation}
\end{theorem}

\begin{proof}
    We start by restating the definition of sensitivity from \cref{eq:sensitivity}:
    \begin{equation}
\resizebox{\linewidth}{!}{$\displaystyle
\text{A-SST} = \frac{1}{N}\sum_{i=1}^N |\Delta (\hat{\mathcal{Y}}^\text{max}_i, \hat{\mathcal{Y}}^\text{min}_i) -\Delta (\hat{\mathcal{Y}}^\text{max}_i, \hat{\mathcal{Y}}^\text{max}_i)|
$}
\end{equation}
    
    \noindent where $\hat{\mathcal{Y}}^\text{max}, \hat{\mathcal{Y}}^\text{min} \in \mathcal{E}(\mathcal{Y})$ are the largest and smallest sets within the partition respectively, with $i$ corresponding to a specific input $\mathbf{x}_i$ in the dataset.
    We treat each of the terms in the difference separately next.
    Firstly, we have 
    \begin{align}
        & \Delta (\hat{\mathcal{Y}}^\text{max}_i, \hat{\mathcal{Y}}^\text{min}_i) \\
        & = \frac{1}{|\hat{\mathcal{Y}}^\text{max}_i||\hat{\mathcal{Y}}^\text{min}_i|}\sum_{\mathbf{y} \in \hat{\mathcal{Y}}^\text{max}_i}\sum_{\mathbf{y}^\prime \in \hat{\mathcal{Y}}^\text{min}_i}\! |c(\mathbf{y})-c(\mathbf{y}^\prime)| \\
        & > \frac{1}{|\hat{\mathcal{Y}}^\text{max}_i||\hat{\mathcal{Y}}^\text{min}_i|}\sum_{\mathbf{y} \in \hat{\mathcal{Y}}^\text{max}_i}\sum_{\mathbf{y}^\prime \in \hat{\mathcal{Y}}^\text{min}_i}\varepsilon = \varepsilon,
    \end{align}
    \noindent where the last step follows from \cref{def:sensitivity}.
    We can also simplify $\Delta (\hat{\mathcal{Y}}^\text{max}_i, \hat{\mathcal{Y}}^\text{max}_i)$ by noticing that from the $|\hat{\mathcal{Y}}^\text{max}_i|^2$ differences considered, $|\hat{\mathcal{Y}}^\text{max}_i|$ of them will amount to $0$ when an element is compared to itself. 
    Using the definition of stability in \cref{def:stability}, we also know that the remaining $|\hat{\mathcal{Y}}^\text{max}_i|^2 - |\hat{\mathcal{Y}}^\text{max}_i|$ comparisons amount to at most $\varepsilon$ each.
    Therefore, 
    \begin{align}
        \Delta (\hat{\mathcal{Y}}^\text{max}_i, \hat{\mathcal{Y}}^\text{max}_i) & \le \frac{1}{|\hat{\mathcal{Y}}^\text{max}_i|^2}\big(|\hat{\mathcal{Y}}^\text{max}_i|^2 - |\hat{\mathcal{Y}}^\text{max}_i|\big)\varepsilon \\
        & = \varepsilon - \frac{1}{|\hat{\mathcal{Y}}^\text{max}_i|}\varepsilon = \frac{|\hat{\mathcal{Y}}^\text{max}_i| - 1}{|\hat{\mathcal{Y}}^\text{max}_i|}\varepsilon.
    \end{align}
    Putting the two arts together, we obtain
    \begin{align}
        \text{A-SST} & > \frac{1}{N}\sum_{i=1}^N \bigg|\varepsilon - \frac{|\hat{\mathcal{Y}}^\text{max}_i| - 1}{|\hat{\mathcal{Y}}^\text{max}_i|}\varepsilon \bigg| \\
        & =  \frac{1}{N}\sum_{i=1}^N  \bigg|\frac{1}{|\hat{\mathcal{Y}}^\text{max}_i|}\varepsilon \bigg| \\
        & = \frac{\varepsilon}{N}\sum_{i=1}^N |\hat{\mathcal{Y}}^\text{max}_i|^{-1}.
    \end{align}
\end{proof}

\vfill

\pagebreak

\section{Additional results}
\subsection{Additional Dataset Results of  Metric Values} \label[appendix]{sec:nq-triviaqa}
Similar to \cref{tab:sciq-popqa}, \cref{tab:nq-triviaqa} presents the results on the NQ and TriviaQA datasets averaged across all models. Overall, we observe trends highly consistent with those of our proposed metrics on SciQ and PopQA.
\begin{table*}[htb]
\centering\footnotesize
\sisetup{round-mode=places, round-precision=2, table-format=1.2, table-number-alignment=center,detect-weight=true,
  detect-family=true}
\setlength{\tabcolsep}{2pt}
\renewcommand{\arraystretch}{1.1}
\resizebox{\linewidth}{!}{%
\begin{tabular}{@{}c l *{6}{S} @{\hspace{1em}} *{6}{S}@{}}
\toprule
& & \multicolumn{6}{c}{\textbf{NQ}} & \multicolumn{6}{c}{\textbf{TriviaQA}}\\
\cmidrule(lr){3-8} \cmidrule(lr){9-14}
& Method
& {ECE$\downarrow$} & {Brier$\downarrow$} & {AUROC$\uparrow$} & {P-RB$\uparrow$} & {A-STB$\uparrow$} & {A-SST$\uparrow$}
& {ECE$\downarrow$} & {Brier$\downarrow$} & {AUROC$\uparrow$} & {P-RB$\uparrow$} & {A-STB$\uparrow$} & {A-SST$\uparrow$} \\
\midrule
\multirow{4}{*}{\rotatebox[origin=c]{90}{\scriptsize Logit-based}}
  & Seq. Likelihood & 0.404 & 0.374 & 0.716 & 0.952 & 0.942 & 0.163 & 0.275 & 0.282 & 0.702 & 0.950 & 0.924 & 0.200 \\
  & Boosted Prob.   & 0.568 & 0.543 & 0.713 & \bfseries 0.988 &  0.988 & 0.033 & 0.383 & 0.371 &\bfseries 0.713 & \bfseries 0.992 &  0.991 & 0.033 \\
  & Platt Scaling   & 0.227 & 0.273 & 0.716 & \bfseries 0.990 & 0.988 & 0.034 & 0.099 & 0.236 & 0.702 & \bfseries 0.989 & 0.983 & 0.045 \\
  & P(True)         & 0.345 & 0.354 & 0.722 & 0.929 & 0.947 & \bfseries 0.201 & 0.268 & 0.283 & 0.700 & 0.947 & 0.955 & \bfseries 0.278 \\
\cmidrule(l){2-14}
\multirow{4}{*}{\rotatebox[origin=c]{90}{\hspace{-0.3cm}\scriptsize Internal States}}
  & Attention Score & \bfseries 0.044 & 0.226 & 0.583 & \bfseries 0.987 &  0.987 & 0.029 & 0.033 & \bfseries 0.234 & 0.558 & \bfseries 0.988 &  0.987 & 0.034 \\
  & Hidden Score    & 0.048 & \bfseries 0.223 & 0.610 & 0.983 & 0.981 & 0.043 & \bfseries 0.021 & 0.235 & 0.521 & \bfseries 0.994 & 0.994 & 0.016 \\
  & SAPLMA   & 0.256 & 0.269 & \bfseries 0.749 & 0.873 & 0.912 & 0.182 & 0.254 & 0.284 & 0.670 & 0.860 & 0.893 & 0.227 \\
  & P(IK)           & 0.202 & 0.266 & 0.642 & 0.959 & \bfseries 1.000 &  0.000 & 0.104 & \bfseries 0.230 & 0.643 & 0.958 & \bfseries 1.000 & 0.000 \\
\cmidrule(l){2-14}
\multirow{1}{*}{\rotatebox[origin=c]{90}{\hspace{1cm}\scriptsize Ling.\hspace{-0.1cm}}}
  & Verbalized Conf.& 0.548 & 0.532 & 0.604 & \bfseries 0.985 & 0.987 & 0.045 & 0.370 & 0.370 & 0.603 & \bfseries 0.985 & 0.987 & 0.082 \\
\cmidrule(l){2-14}
\multirow{1}{*}{\rotatebox[origin=l]{90}{\scriptsize\shortstack{Aux.}\hspace{-0.1cm}}}
  & Calib1          & 0.064 & \bfseries 0.216 & 0.658 & 0.983 & 0.972 & 0.068 & 0.065 & \bfseries 0.232 & 0.595 & \bfseries 0.989 & 0.978 & 0.055 \\
\bottomrule
\end{tabular}}
\caption{Calibration and our metrics averaged across all models. Arrows indicate metric direction ($\downarrow$ lower is better, $\uparrow$ higher is better); the best value per column is bold.}\label{tab:nq-triviaqa}
\end{table*}

\subsection{Judge Reliability}

To validate judge reliability, we conducted a small-scale study involving two human annotators (Human 1 with an AI background and Human 2 without an AI background) and GPT-5.5 as an independent judge. We evaluated 200 sampled correctness and semantic-partition judgments, with 50 examples from each dataset. All judges are provided with the same prompts as in the paper. 

The results are shown in Table \ref{tab:judge-agreement}. For correctness judgments, GPT-4o shows strong agreement with both human annotators and GPT-5.5. Agreement across all four judges is also high, with a Krippendorff's alpha of 0.814. GPT-4o's agreement with the human-consensus labels is nearly identical to that of GPT-5.5, suggesting that the correctness labels are not artifacts of the original GPT-4o judge.

For semantic partitioning, we conducted an acceptability audit to determine whether the independent validators approved the partitions generated by GPT-4o. The results are shown in Table \ref{tab:partition-approval}. Approval rates are high for both human annotators and GPT-5.5.

\begin{table}[t]
\centering
\small
\setlength{\tabcolsep}{4.8pt}
\renewcommand{\arraystretch}{1}
\begin{tabular}{lcc}
\toprule
\textbf{Comparison} & \textbf{Agreement} & \textbf{Cohen's $\kappa$} \\
\midrule
\multicolumn{3}{l}{\textit{Human agreement}} \\
Human 1 vs.\ Human 2
    & 0.930 & 0.839 \\
\midrule

\multicolumn{3}{l}{\textit{Model--human agreement}} \\
GPT-4o vs.\ Human 1
    & 0.930 & 0.842 \\
GPT-4o vs.\ Human 2
    & 0.900 & 0.774 \\
GPT-5.5 vs.\ Human 1
    & 0.930 & 0.832 \\
GPT-5.5 vs.\ Human 2
    & 0.910 & 0.784 \\
\midrule

\multicolumn{3}{l}{\textit{Model--model agreement}} \\
GPT-4o vs.\ GPT-5.5
    & 0.920 & 0.812 \\
\midrule

\multicolumn{3}{l}{\textit{Aggregate agreement}} \\
GPT-4o vs.\ human consensus
    & 0.946 & -- \\
GPT-5.5 vs.\ human consensus
    & \textbf{0.952} & -- \\
All judges
    & 0.855 & -- \\
\bottomrule
\end{tabular}
\caption{Agreement on response-correctness judgments among human and
LLM evaluators. Human consensus is defined over items for which the two
human annotators agreed. All-judge agreement is the proportion of items
on which all four evaluators assigned the same label. Cohen's $\kappa$
is reported only for pairwise comparisons over the full sample.}
\label{tab:judge-agreement}
\end{table}

\begin{table}[t]
\centering
\small
\begin{tabular}{lc}
\toprule
\textbf{Validator} & \textbf{Approval Rate} \\
\midrule
Human 1        & 0.975 \\
Human 2        & 0.965 \\
GPT-5.5        & 0.905 \\
All validators & 0.884 \\
\bottomrule
\end{tabular}
\caption{Validator approval rates for the generated answer partitions.}
\label{tab:partition-approval}
\end{table}
\subsection{Accuracy of Models}

\begin{figure*}[hbp]
    \centering
    \includegraphics[width=1\linewidth]{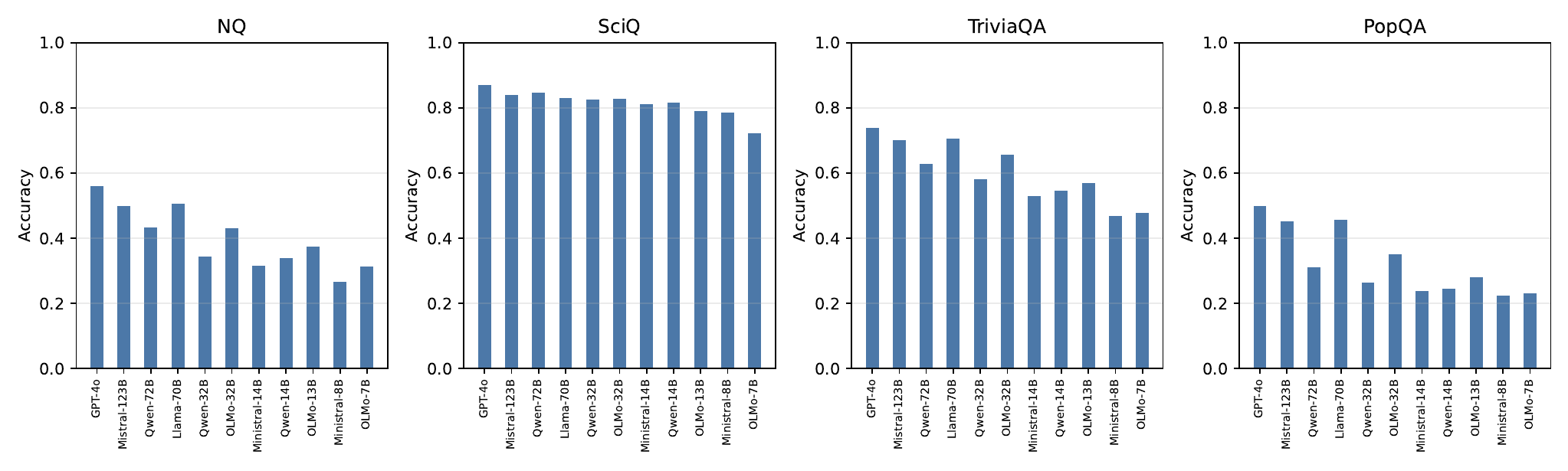}
    \caption{Accuracy of models on different datasets.}
    \label{fig:acc}
\end{figure*}

We present the detailed model accuracy on each benchmark dataset in \cref{fig:acc}. Overall, all models achieve the strongest performance on SciQ and the weakest on PopQA, with the performance gap between the two datasets reaching nearly 50\%. This highlights the substantial diversity and varying difficulty of the evaluated benchmarks.

\subsection{Distribution of Metric Values}
\begin{figure*}[hbp]
    \centering
    \includegraphics[width=0.95\linewidth]{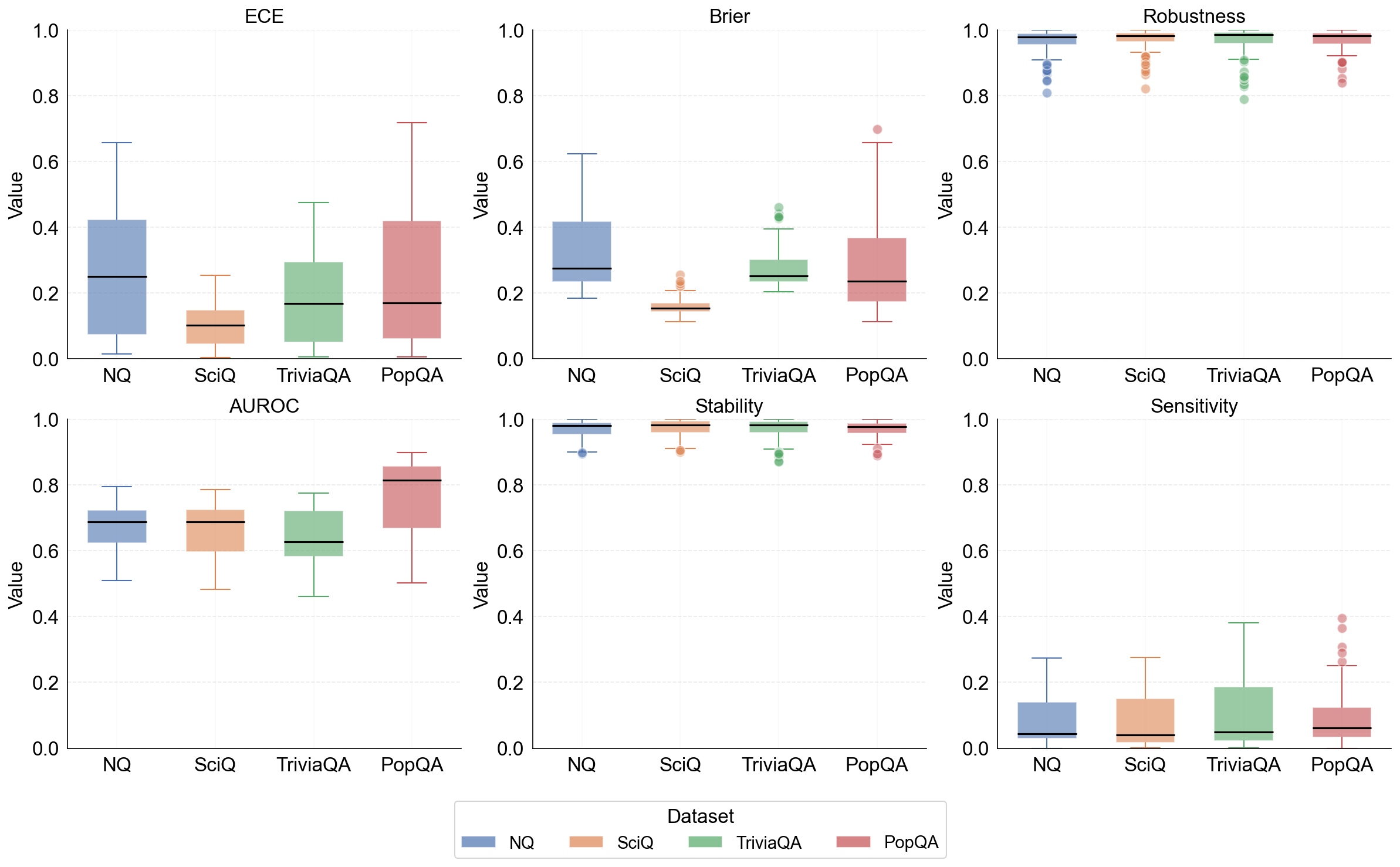}
    \caption{Distribution of metric across models and CE methods, shown per-dataset. Whiskers show the interquartile range and dots signify outliers.}
    \label{fig:boxplots-datasets}
\end{figure*}

\label[appendix]{sec:distribution-add}
\cref{fig:boxplots-datasets} shows the distribution of metric values across models and CE methods for each dataset. The results demonstrate that CE performance varies substantially across datasets, with no single dataset consistently yielding higher metric values across all metrics. Moreover, each metric exhibits distinct distributional characteristics: robustness and stability are consistently skewed toward high values, whereas sensitivity remains uniformly low across datasets. At the same time, the distributions of our proposed metrics differ markedly from those of ECE, AUROC, and Brier score, further suggesting that they capture complementary aspects of confidence quality.

\subsection{Statistic Test} \label[appendix]{sec:significance}
We report repeated-measures-aware Kendall's $\tau$ correlations between the evaluation metrics in Table \ref{tab:metric-correlations}. Each cell presents the mean dataset-wise tau with a cluster-bootstrap 95\% confidence interval. Stars indicate Benjamini-Hochberg-corrected permutation p-values: * (p < 0.05), ** (p < 0.01), *** (p < 0.001).

The corrected analysis supports our original conclusion: ECE and Brier are strongly related, whereas P-RB, A-STB, and A-SST capture complementary dimensions.

\begin{table*}[t]
\centering
\small
\setlength{\tabcolsep}{4pt}
\renewcommand{\arraystretch}{1.25}
\begin{tabular}{lccccccc}
\toprule
\textbf{Metric}
& \textbf{ECE}
& \textbf{Brier}
& \textbf{AUROC}
& \textbf{P-RB}
& \textbf{A-STB}
& \textbf{A-SST}
& \textbf{Acc.} \\
\midrule

ECE
& 1 & & & & & & \\

Brier
& $0.644^{***}$
& 1 & & & & & \\
& {\scriptsize [0.590, 0.696]} \\

AUROC
& $0.303^{***}$
& $0.084^{*}$
& 1 & & & & \\
& {\scriptsize [0.234, 0.368]}
& {\scriptsize [0.019, 0.157]} \\

P-RB
& $-0.102^{**}$
& $0.016$
& $-0.330^{***}$
& 1 & & & \\
& {\scriptsize [$-0.168$, $-0.032$]}
& {\scriptsize [$-0.049$, 0.078]}
& {\scriptsize [$-0.390$, $-0.270$]} \\

A-STB
& $-0.148^{***}$
& $-0.061$
& $-0.249^{***}$
& $0.502^{***}$
& 1 & & \\
& {\scriptsize [$-0.205$, $-0.090$]}
& {\scriptsize [$-0.123$, 0.001]}
& {\scriptsize [$-0.309$, $-0.190$]}
& {\scriptsize [0.436, 0.569]} \\

A-SST
& $0.214^{***}$
& $0.120^{***}$
& $0.262^{***}$
& $-0.466^{***}$
& $-0.795^{***}$
& 1 & \\
& {\scriptsize [0.155, 0.273]}
& {\scriptsize [0.058, 0.181]}
& {\scriptsize [0.200, 0.325]}
& {\scriptsize [$-0.531$, $-0.402$]}
& {\scriptsize [$-0.824$, $-0.764$]} \\

Acc.
& $-0.065$
& $-0.169^{***}$
& $-0.162^{***}$
& $0.045$
& $0.023$
& $0.037$
& 1 \\
& {\scriptsize [$-0.136$, 0.004]}
& {\scriptsize [$-0.232$, $-0.104$]}
& {\scriptsize [$-0.230$, $-0.093$]}
& {\scriptsize [$-0.025$, 0.115]}
& {\scriptsize [$-0.044$, 0.089]}
& {\scriptsize [$-0.035$, 0.109]} \\
\bottomrule
\end{tabular}

\caption{Mean dataset-wise Kendall's $\tau$ correlations among evaluation
metrics and model accuracy. Brackets report cluster-bootstrap 95\%
confidence intervals. Asterisks indicate BH-corrected permutation
significance: $^{*}p<0.05$, $^{**}p<0.01$, and
$^{***}p<0.001$.}
\label{tab:metric-correlations}
\end{table*}

\subsection{Retained Fraction of Prompts for Prompt Robustness}

We report the average retained fraction $|I|/M$
in Table \ref{tab:retained-fraction}, where $I$
 is the subset of prompts producing answers semantically equivalent to the answer generated from the default prompt. The higher-accuracy datasets, SciQ and TriviaQA, have the highest retained fractions. Nevertheless, all datasets also retain substantial proportions of prompts. 

 \begin{table}[t]
\centering
\small
\begin{tabular}{lcccc}
\toprule
& \textbf{NQ} & \textbf{PopQA} & \textbf{SciQ} & \textbf{TriviaQA} \\
\midrule
$|{I}|/{M}$ & 0.698 & 0.643 & 0.860 & 0.877 \\
\bottomrule
\end{tabular}
\caption{Average fraction of prompts retained $|{I}|/{M}$ after semantic-equivalence
filtering.}
\label{tab:retained-fraction}
\end{table}
\subsection{Detailed Scaling Results} 
\begin{figure*}[thb]
    \centering

    \begin{subfigure}[t]{0.95\linewidth}
        \centering
        \includegraphics[width=\textwidth]{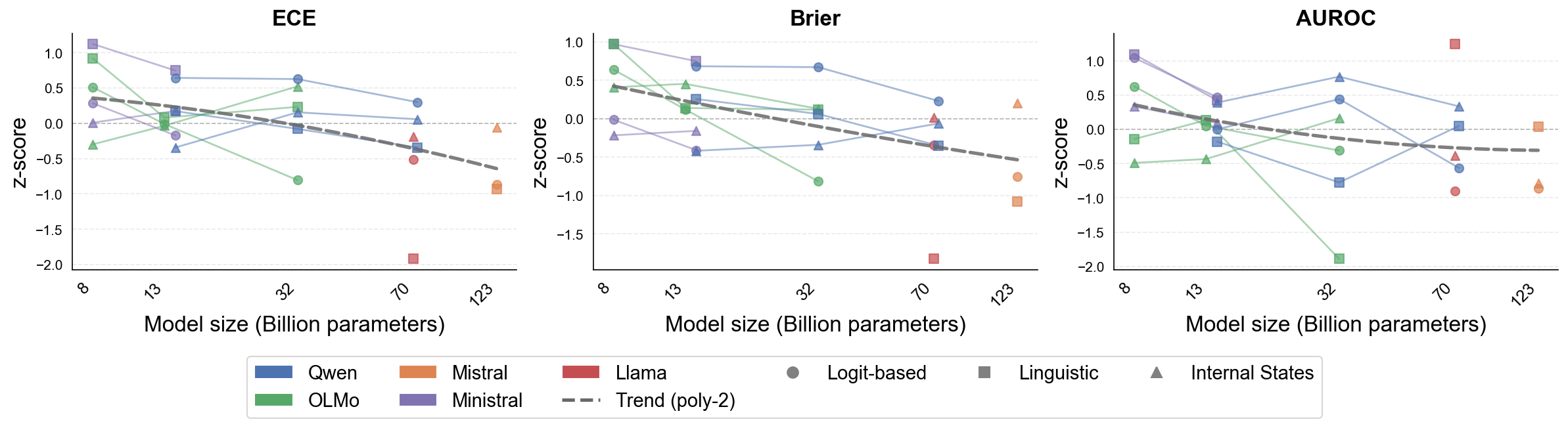}
    \end{subfigure}
    ~
    \begin{subfigure}[t]{0.95\linewidth}
        \centering
        \includegraphics[width=\textwidth]{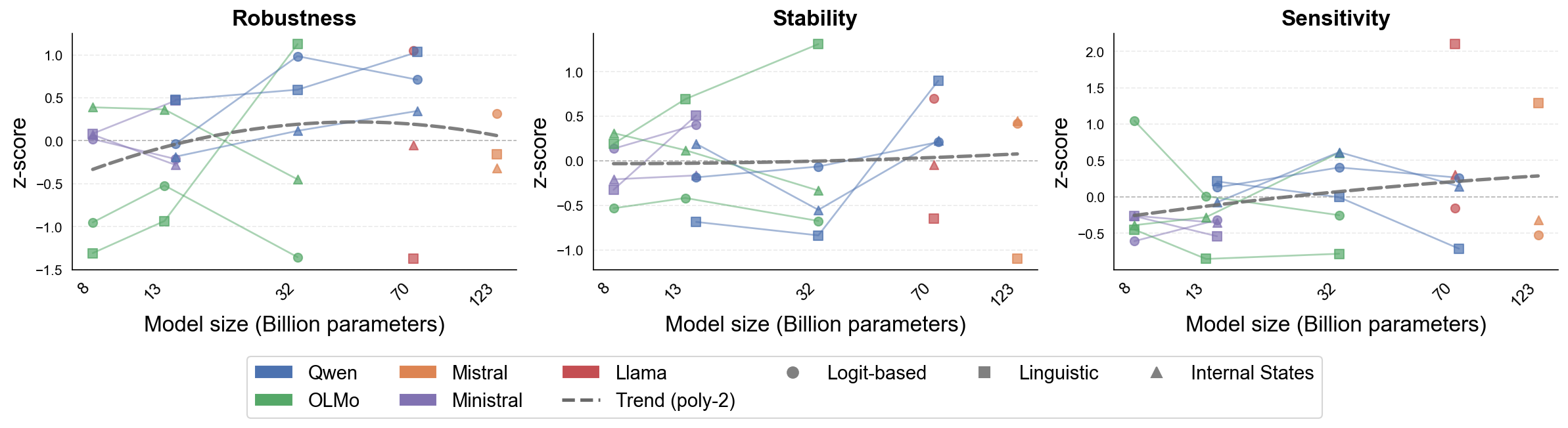}
    \end{subfigure}
    \caption{Normalized metric value distributions by model family and size. 
    Confidence estimators from the same estimator family and for the same model family are connected and appear in the same color. 
    Measurements from confidence estimators in the same family are shown with the same marker.
    Trend lines are second-degree polynomials fitted on all observations.}\label{fig:scaling-laws-method-families}
\end{figure*}

\label[appendix]{sec:scaling-add}
We present more detailed scaling results in \cref{fig:scaling-laws-method-families}. For linguistic CE methods, such as verbalized confidence, scaling generally improves robustness and stability, while sensitivity exhibits inconsistent trends. In contrast, the trends for the other CE method families are substantially less consistent across metrics and model scales. Overall, the results suggest that the effect of scaling strongly depends on both the CE method type and the underlying model family.

\subsection{Relationship between A-SST and answer correctness.}
We additionally examine whether A-SST reflects differences in answer
correctness. For each example, we compare representative answers from
the largest and smallest semantic groups, label the correctness of each
answer, and separate cases based on whether the two answers have the same
or different correctness labels.

Averaged across datasets, A-SST increases most clearly for answer-sensitive
methods when the semantic groups also differ in correctness, as shown in
\cref{tab:asst-correctness}.

\begin{table}[t]
\centering
\footnotesize
\setlength{\tabcolsep}{4pt}
\begin{tabular}{lccc}
\toprule
\textbf{Method}
& \shortstack{$\text{A-SST}_{\mathrm{same}}$}
& \shortstack{$\text{A-SST}_{\mathrm{different}}$}
& \textbf{$\Delta$} \\
\midrule
Seq.\ Likelihood  & 0.183 & 0.200 &  0.017 \\
Boosted Prob.     & 0.029 & 0.014 & $-0.015$ \\
Platt Scaling     & 0.039 & 0.043 &  0.004 \\
P(True)           & 0.231 & 0.347 &  \textbf{0.116} \\
Attention Score   & 0.028 & 0.033 &  0.005 \\
Hidden Score      & 0.040 & 0.046 &  0.006 \\
SAPLMA             & 0.170 & 0.270 &  \textbf{0.100} \\
P(IK)             & 0.000 & 0.000 &  0.000 \\
Verbalized Conf.  & 0.084 & 0.089 &  0.006 \\
Calib1            & 0.067 & 0.083 &  0.016 \\
\bottomrule
\end{tabular}
\caption{Mean A-SST for pairs of semantic groups with the same $\text{A-SST}_{\mathrm{same}}$ or
different $\text{A-SST}_{\mathrm{different}}$ answer-correctness labels. The difference is calculated as
$\Delta=\text{A-SST}_{\mathrm{different}}
-\text{A-SST}_{\mathrm{same}}$.}
\label{tab:asst-correctness}
\end{table}

The largest increases are observed for P(True) ($\Delta=0.116$) and
SAPLMA ($\Delta=0.100$), both of which use information from the generated
answer. In contrast, P(IK), which depends only on the question, shows no
change ($\Delta=0.000$).

These results provide supporting evidence for the intended interpretation
of A-SST: for some answer-sensitive methods, confidence changes more when
semantic answer differences coincide with differences in correctness.

\begin{figure*}[thb]
    \centering
    \includegraphics[width=\textwidth]{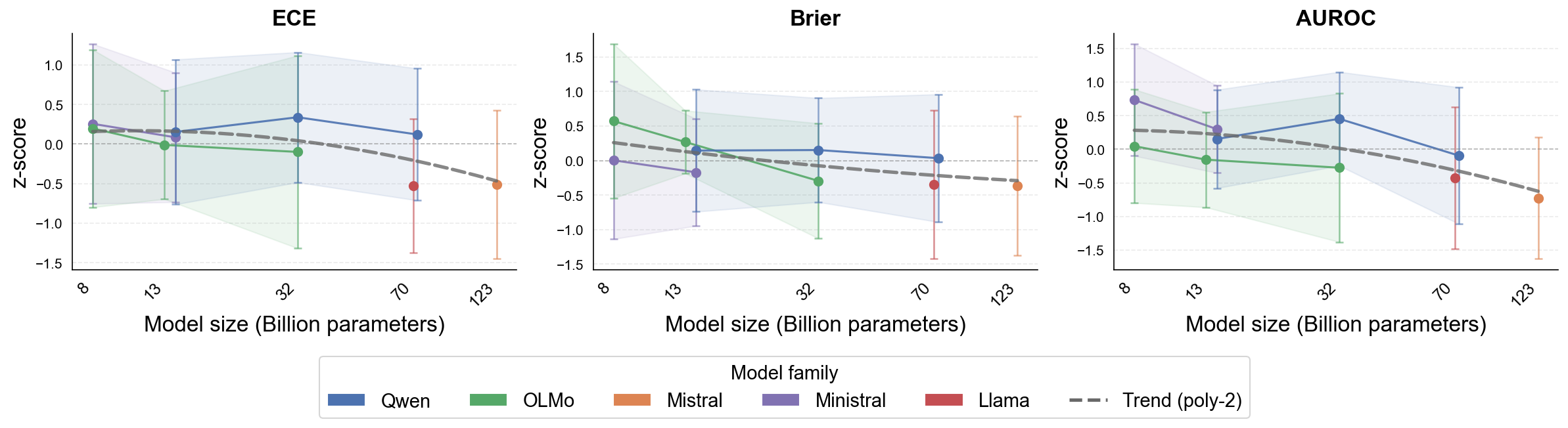}
    \caption{Normalized metric value distributions by model family and size. 
    Whiskers and shaded regions indicate the standard deviation over observed metric values across datasets and CE methods. 
    Models of the same family are connected and appear in the same color. Trend lines are  second degree polynomials fitted on all observations.}\label{fig:scaling-laws-shaded-calib}
\end{figure*}

\subsection{Detailed Reciprocal Rank of CE Methods} \label[appendix]{sec:rank-add}

\begin{figure*}
    \centering
    \includegraphics[width=0.9\textwidth]{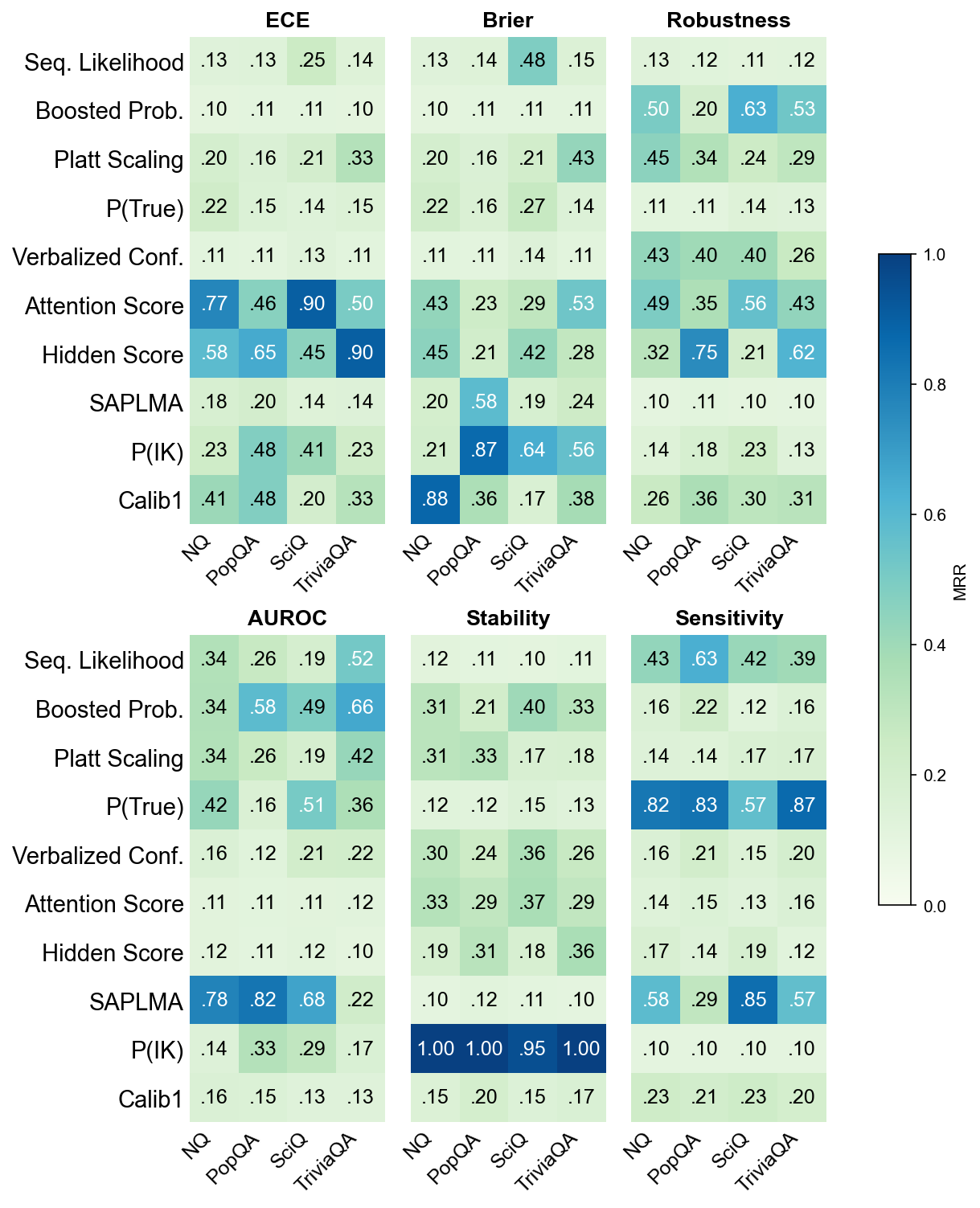}
    \caption{Reciprocal rank of CE methods per evaluation metric over datasets.}\label{fig:mrr-per-dataset}
\end{figure*}

The overall mean reciprocal rank (MRR) of different CE methods for each dataset is shown in \cref{fig:mrr-per-dataset}. We observe that CE method performance varies across datasets, and that no single method consistently performs best across all datasets and metrics.

Compared to ECE, Brier score, and AUROC, the cross-dataset variation is smaller for robustness, stability, and sensitivity. We hypothesize that this is because traditional calibration and discrimination metrics are more directly tied to correctness, and are therefore more strongly affected by dataset difficulty.

\subsection{Case Study} \label[appendix]{sec:case-study}

\paragraph{Robustness.} We show two robustness examples in \cref{tab:case-robust-detail} and \cref{tab:case-robust-detail-19}. The results indicate that, across all prompts ($t_1$--$t_{10}$), the Qwen-14B and Mistral models generate semantically equivalent answers. Although the confidence scores produced by different CE methods vary, most methods remain robust to prompt perturbations. In contrast, P(True) and Verbalized Conf. are noticeably less robust in these two cases (P-RB = 0.5), consistent with the trend observed in the main paper. We hypothesize that methods relying more heavily on prompt formatting are more impacted by prompt perturbations. 

\paragraph{Stability and Sensitivity.} 
We present three qualitative examples illustrating stability and sensitivity in \cref{tab:case-stab-sens-detail}-\cref{tab:case-stab-sens-detail-19}. For stability, SAPLMA is the least stable method across semantically equivalent answers. We hypothesize that this is because SAPLMA is trained on the hidden state of the final generated token; therefore, small lexical changes at the end of the response can substantially alter the estimated confidence. For example, in \cref{tab:case-stab-sens-detail}, the semantically equivalent answers ``Concave lens'' and ``Concave'' end with different tokens, resulting in confidence scores of 0.07 and 1.00, respectively.

For sensitivity, P(True), which directly leverages generation-side information, performs best, followed by SAPLMA, which focuses on the final generated token. Methods such as sequence likelihood also use generation-side information, but since all candidate answers remain highly likely generations, the resulting confidence differences are relatively small. Post-hoc scaling methods, such as Platt scaling, further compress these differences and therefore reduce sensitivity. In contrast, methods that do not use answer-side information, such as P(IK), which is trained only on the hidden states of the final token of the question, are almost completely insensitive to answer changes (A-SST = 0). Although Calib1 conditions on both the question and answer, the short-form nature of our datasets likely reduces the contribution of the answer representation, resulting in moderate but still limited sensitivity.

\begin{table*}[htb]
\centering
\footnotesize
\setlength{\tabcolsep}{6pt}
\begin{tabular}{lllrrrrrrrrrrr}
\toprule
\multicolumn{14}{c}{The electrode at which oxidation occurs is called?  ~$y$: the anode. ~Corr.: 1.00} \\
\midrule
\textbf{}& \textbf{} & \textbf{CE}& \textbf{$t_1$} & $t_2$  & $t_3$  & $t_4$ & $t_5$ & $t_6$  & $t_7$  & $t_8$  &$t_9$ &$t_{10}$ & \textbf{P-RB}\\
\cmidrule(lr){1-3} \cmidrule(lr){4-13} \cmidrule(lr){14-14}
\multirow{10}{*}{ \rotatebox[origin=c]{90}{Qwen-14B} } & \multirow{10}{*}{ \rotatebox[origin=c]{90}{SciQ} } & Seq. Likelihood & 0.96 & 0.72 & 0.98 & 1.00 & 1.00 & 1.00 & 0.99 & 0.89 & 1.00 & 1.00 & 0.92 \\
 &  & Boosted Prob. & 1.00 & 1.00 & 1.00 & 1.00 & 1.00 & 1.00 & 1.00 & 1.00 & 1.00 & 1.00 & 1.00 \\
 &  & Platt Scaling & 0.75 & 0.75 & 0.75 & 0.75 & 0.75 & 0.75 & 0.75 & 0.73 & 0.75 & 0.75 & 0.99 \\
 &  & P(True) & 1.00 & 1.00 & 0.00 & 0.00 & 1.00 & 1.00 & 0.00 & 1.00 & 0.00 & 0.00 & 0.50 \\
 &  & Attention Score & 0.81 & 0.81 & 0.80 & 0.81 & 0.81 & 0.81 & 0.80 & 0.80 & 0.81 & 0.81 & 1.00 \\
 &  & Hidden Score & 0.85 & 0.86 & 0.84 & 0.84 & 0.83 & 0.86 & 0.84 & 0.85 & 0.83 & 0.83 & 0.99 \\
 &  & SAPLMA & 1.00 & 1.00 & 1.00 & 0.99 & 0.99 & 1.00 & 1.00 & 1.00 & 0.99 & 0.98 & 0.99 \\
 &  & P(IK) & 0.90 & 0.89 & 0.88 & 0.91 & 0.91 & 0.91 & 0.89 & 0.89 & 0.93 & 0.95 & 0.98 \\
 &  & Verbalized Conf. & 0.95 & 0.95 & 0.95 & 0.95 & 0.95 & 0.95 & 0.95 & 0.95 & 0.95 & 0.95 & 1.00 \\
 &  & Calib1 & 0.77 & 0.76 & 0.77 & 0.76 & 0.76 & 0.78 & 0.77 & 0.75 & 0.74 & 0.76 & 0.99 \\
\bottomrule
\end{tabular}
\caption{Case study 1 of prompt robustness. The confidence estimated by different CE methods under 10 perturbed answer-elicit prompts ($t_1$ -- $t_{10}$) for the same question and LLM response is shown. }
\label{tab:case-robust-detail}
\end{table*}

\begin{table*}
\centering
\footnotesize
\setlength{\tabcolsep}{5pt}
\begin{tabular}{llcrrrrrrrrrrr}
\toprule
\multicolumn{14}{c}{With an atomic weight of 22, what element, named for members of Greek mythology, uses the symbol Ti? } \\
\multicolumn{14}{c}{ ~$y$: Titanium ore. ~Corr.: 1.00} \\
\midrule
\textbf{}& \textbf{} & \textbf{CE}& \textbf{$t_1$} & $t_2$  & $t_3$  & $t_4$ & $t_5$ & $t_6$  & $t_7$  & $t_8$  &$t_9$ &$t_{10}$ & \textbf{P-RB}\\
\cmidrule(lr){1-3} \cmidrule(lr){4-13} \cmidrule(lr){14-14}
\multirow{10}{*}{ \rotatebox[origin=c]{90}{Mistral-123B} } & \multirow{10}{*}{ \rotatebox[origin=c]{90}{TriviaQA} } & Seq. Likelihood & 0.99 & 0.87 & 0.90 & 0.92 & 0.79 & 1.00 & 0.91 & 0.99 & 0.90 & 0.84 & 0.93 \\
 &  & Boosted Prob. & 1.00 & 0.99 & 1.00 & 0.99 & 0.97 & 1.00 & 1.00 & 1.00 & 0.99 & 0.99 & 0.99 \\
 &  & Platt Scaling & 0.73 & 0.70 & 0.71 & 0.71 & 0.68 & 0.73 & 0.71 & 0.73 & 0.71 & 0.70 & 0.99 \\
 &  & P(True) & 0.08 & 0.01 & 0.05 & 0.01 & 0.65 & 0.08 & 0.03 & 0.08 & 0.02 & 0.84 & 0.72 \\
 &  & Attention Score & 0.69 & 0.71 & 0.73 & 0.71 & 0.70 & 0.69 & 0.71 & 0.69 & 0.71 & 0.71 & 0.99 \\
 &  & Hidden Score & 0.72 & 0.76 & 0.76 & 0.76 & 0.76 & 0.72 & 0.76 & 0.73 & 0.76 & 0.76 & 0.98 \\
 &  & SAPLMA & 0.98 & 0.22 & 0.00 & 0.30 & 0.07 & 0.96 & 0.02 & 0.94 & 0.38 & 0.13 & 0.62 \\
 &  & P(IK) & 0.70 & 0.51 & 0.45 & 0.70 & 0.67 & 0.69 & 0.52 & 0.42 & 0.76 & 0.57 & 0.89 \\
 &  & Verbalized Conf. & 1.00 & 0.00 & 0.00 & 0.00 & 1.00 & 1.00 & 0.00 & 1.00 & 0.00 & 1.00 & 0.50 \\
 &  & Calib1 & 0.61 & 0.61 & 0.61 & 0.61 & 0.61 & 0.61 & 0.61 & 0.61 & 0.61 & 0.61 & 1.00 \\
\bottomrule
\end{tabular}
\caption{Case study 2 of prompt robustness. The confidence estimated by different CE methods under 10 perturbed prompts $t_1$ -- $t_{10}$)  for the same question and LLM response is shown.}
\label{tab:case-robust-detail-19}
\end{table*}

\begin{table*}
\centering
\footnotesize
\setlength{\tabcolsep}{8pt}
\begin{tabular}{lllrrrrr}
\toprule
\multicolumn{8}{c}{What type of lens is thicker at the edges than it is in the middle? ~~ $y^*$: concave lens} \\
 &  &  & \cellcolor[HTML]{E6F4D7}Convex & \cellcolor[HTML]{EADCF8}Concave lens & \cellcolor[HTML]{EADCF8}Concave &  &  \\
\midrule
& & CE & \cellcolor[HTML]{E6F4D7}$\hat{\mathcal{Y}}_{1}$ & \cellcolor[HTML]{EADCF8}$\hat{\mathcal{Y}}_{2}$ & \cellcolor[HTML]{EADCF8}$\hat{\mathcal{Y}}_{2}$ & \textbf{A-STB} & \textbf{A-SST} \\
\cmidrule(lr){1-3} \cmidrule(lr){4-6} \cmidrule(lr){7-8}
\multirow{10}{*}{\rotatebox[origin=c]{90}{Llama-70B} } & \multirow{10}{*}{ \rotatebox[origin=c]{90}{SciQ} } & Seq. Likelihood & \cellcolor[HTML]{E6F4D7}0.59 & \cellcolor[HTML]{EADCF8}0.59 & \cellcolor[HTML]{EADCF8}0.82 & 0.90 & 0.17 \\
 &  & Boosted Prob. & \cellcolor[HTML]{E6F4D7}1.00 & \cellcolor[HTML]{EADCF8}1.00 & \cellcolor[HTML]{EADCF8}1.00 & 1.00 & 0.00 \\
 &  & Platt Scaling & \cellcolor[HTML]{E6F4D7}0.67 & \cellcolor[HTML]{EADCF8}0.67 & \cellcolor[HTML]{EADCF8}0.73 & 0.98 & 0.04 \\
 &  & P(True) & \cellcolor[HTML]{E6F4D7}0.32 & \cellcolor[HTML]{EADCF8}0.00 & \cellcolor[HTML]{EADCF8}0.00 & 1.00 & 0.32 \\
 &  & Attention Score & \cellcolor[HTML]{E6F4D7}0.84 & \cellcolor[HTML]{EADCF8}0.84 & \cellcolor[HTML]{EADCF8}0.84 & 1.00 & 0.00 \\
 &  & Hidden Score & \cellcolor[HTML]{E6F4D7}0.90 & \cellcolor[HTML]{EADCF8}0.77 & \cellcolor[HTML]{EADCF8}0.78 & 1.00 & 0.11 \\
 &  & SAPLMA & \cellcolor[HTML]{E6F4D7}1.00 & \cellcolor[HTML]{EADCF8}0.07 & \cellcolor[HTML]{EADCF8}1.00 & 0.60 & 0.23 \\
 &  & P(IK) & \cellcolor[HTML]{E6F4D7}0.90 & \cellcolor[HTML]{EADCF8}0.90 & \cellcolor[HTML]{EADCF8}0.90 & 1.00 & 0.00 \\
 &  & Verbalized Conf. & \cellcolor[HTML]{E6F4D7}1.00 & \cellcolor[HTML]{EADCF8}0.00 & \cellcolor[HTML]{EADCF8}0.00 & 1.00 & 1.00 \\
 &  & Calib1 & \cellcolor[HTML]{E6F4D7}0.87 & \cellcolor[HTML]{EADCF8}0.81 & \cellcolor[HTML]{EADCF8}0.78 & 0.99 & 0.09 \\
\bottomrule
\end{tabular}
\caption{Case study 1 of answer stability and sensitivity. The confidence estimated by different CE methods for unique sampled answers to the same question is shown. Purple marks the answer group used for A-STB; purple and green mark the semantic groups used for A-SST.}
\label{tab:case-stab-sens-detail}
\end{table*}

\begin{table*}
\centering
\footnotesize
\setlength{\tabcolsep}{1pt}
\begin{tabular}{lllrrrrrr}
\toprule
\multicolumn{9}{c}{where was the world chess tournament 2017 held? ~~ $y^*$: Tbilisi, Georgia} \\
 &  &  & \cellcolor[HTML]{EADCF8}New York City, USA & Riyadh, Saudi Arabia & \cellcolor[HTML]{EADCF8}New York City & \cellcolor[HTML]{E6F4D7}Carlsen vs. Caruana 2018 &  &  \\
\midrule
& & CE & \cellcolor[HTML]{EADCF8}$\hat{\mathcal{Y}}_{1}$ & $\hat{\mathcal{Y}}_{2}$ & \cellcolor[HTML]{EADCF8}$\hat{\mathcal{Y}}_{3}$ & \cellcolor[HTML]{E6F4D7}$\hat{\mathcal{Y}}_{4}$ & \textbf{A-STB} & \textbf{A-SST} \\
\cmidrule(lr){1-3} \cmidrule(lr){4-7} \cmidrule(lr){8-9}
\multirow{10}{*}{\rotatebox[origin=c]{90}{Mistral-123B} } & \multirow{10}{*}{ \rotatebox[origin=c]{90}{NQ} } & Seq. Likelihood & \cellcolor[HTML]{EADCF8}0.61 & 0.69 & \cellcolor[HTML]{EADCF8}0.74 & \cellcolor[HTML]{E6F4D7}0.64 & 0.94 & 0.03 \\
 &  & Boosted Prob. & \cellcolor[HTML]{EADCF8}0.99 & 0.99 & \cellcolor[HTML]{EADCF8}0.99 & \cellcolor[HTML]{E6F4D7}0.98 & 1.00 & 0.00 \\
 &  & Platt Scaling & \cellcolor[HTML]{EADCF8}0.60 & 0.62 & \cellcolor[HTML]{EADCF8}0.63 & \cellcolor[HTML]{E6F4D7}0.61 & 0.99 & 0.01 \\
 &  & P(True) & \cellcolor[HTML]{EADCF8}0.05 & 0.87 & \cellcolor[HTML]{EADCF8}0.04 & \cellcolor[HTML]{E6F4D7}0.01 & 0.99 & 0.03 \\
 &  & Attention Score & \cellcolor[HTML]{EADCF8}0.53 & 0.53 & \cellcolor[HTML]{EADCF8}0.53 & \cellcolor[HTML]{E6F4D7}0.53 & 1.00 & 0.00 \\
 &  & Hidden Score & \cellcolor[HTML]{EADCF8}0.47 & 0.52 & \cellcolor[HTML]{EADCF8}0.52 & \cellcolor[HTML]{E6F4D7}0.47 & 0.98 & 0.02 \\
 &  & SAPLMA & \cellcolor[HTML]{EADCF8}0.39 & 0.01 & \cellcolor[HTML]{EADCF8}0.97 & \cellcolor[HTML]{E6F4D7}0.00 & 0.74 & 0.57 \\
 &  & P(IK) & \cellcolor[HTML]{EADCF8}0.79 & 0.79 & \cellcolor[HTML]{EADCF8}0.80 & \cellcolor[HTML]{E6F4D7}0.80 & 1.00 & 0.00 \\
 &  & Verbalized Conf. & \cellcolor[HTML]{EADCF8}1.00 & 1.00 & \cellcolor[HTML]{EADCF8}1.00 & \cellcolor[HTML]{E6F4D7}0.00 & 1.00 & 1.00 \\
 &  & Calib1 & \cellcolor[HTML]{EADCF8}0.59 & 0.59 & \cellcolor[HTML]{EADCF8}0.56 & \cellcolor[HTML]{E6F4D7}0.41 & 0.99 & 0.15 \\
\bottomrule
\end{tabular}
\caption{Case study 2 of answer stability and sensitivity. Purple marks the answer group used for A-STB; purple and green mark the semantic groups used for A-SST.}
\label{tab:case-stab-sens-detail-3}
\end{table*}

\begin{table*}
\centering
\footnotesize
\setlength{\tabcolsep}{10pt}
\begin{tabular}{lllrrrrr}
\toprule
\multicolumn{8}{c}{What is the capital of Cao Wei? ~~ $y^*$: Luoyang} \\
 &  &  & \cellcolor[HTML]{EADCF8}{\begin{CJK*}{UTF8}{gbsn}洛阳\end{CJK*}} & \cellcolor[HTML]{E6F4D7}Luo Yin & \cellcolor[HTML]{EADCF8}Luoyang &  &  \\
\midrule
& & CE & \cellcolor[HTML]{EADCF8}$y_{1}$ & \cellcolor[HTML]{E6F4D7}$y_{2}$ & \cellcolor[HTML]{EADCF8}$y_{3}$ & \textbf{A-STB} & \textbf{A-SST} \\
\cmidrule(lr){1-3} \cmidrule(lr){4-6} \cmidrule(lr){7-8}
\multirow{10}{*}{\rotatebox[origin=c]{90}{Qwen-14B} } & \multirow{10}{*}{ \rotatebox[origin=c]{90}{PopQA} } & Seq. Likelihood & \cellcolor[HTML]{EADCF8}0.47 & \cellcolor[HTML]{E6F4D7}0.46 & \cellcolor[HTML]{EADCF8}0.61 & 0.94 & 0.01 \\
 &  & Boosted Prob. & \cellcolor[HTML]{EADCF8}1.00 & \cellcolor[HTML]{E6F4D7}0.97 & \cellcolor[HTML]{EADCF8}1.00 & 1.00 & 0.03 \\
 &  & Platt Scaling & \cellcolor[HTML]{EADCF8}0.49 & \cellcolor[HTML]{E6F4D7}0.49 & \cellcolor[HTML]{EADCF8}0.52 & 0.99 & 0.00 \\
 &  & P(True) & \cellcolor[HTML]{EADCF8}1.00 & \cellcolor[HTML]{E6F4D7}0.00 & \cellcolor[HTML]{EADCF8}1.00 & 1.00 & 1.00 \\
 &  & Attention Score & \cellcolor[HTML]{EADCF8}0.36 & \cellcolor[HTML]{E6F4D7}0.24 & \cellcolor[HTML]{EADCF8}0.18 & 0.92 & 0.03 \\
 &  & Hidden Score & \cellcolor[HTML]{EADCF8}0.39 & \cellcolor[HTML]{E6F4D7}0.16 & \cellcolor[HTML]{EADCF8}0.18 & 0.91 & 0.09 \\
 &  & SAPLMA & \cellcolor[HTML]{EADCF8}0.00 & \cellcolor[HTML]{E6F4D7}0.40 & \cellcolor[HTML]{EADCF8}0.92 & 0.59 & 0.06 \\
 &  & P(IK) & \cellcolor[HTML]{EADCF8}0.23 & \cellcolor[HTML]{E6F4D7}0.23 & \cellcolor[HTML]{EADCF8}0.23 & 1.00 & 0.00 \\
 &  & Verbalized Conf. & \cellcolor[HTML]{EADCF8}0.95 & \cellcolor[HTML]{E6F4D7}0.85 & \cellcolor[HTML]{EADCF8}0.95 & 1.00 & 0.10 \\
 &  & Calib1 & \cellcolor[HTML]{EADCF8}0.49 & \cellcolor[HTML]{E6F4D7}0.21 & \cellcolor[HTML]{EADCF8}0.57 & 0.96 & 0.27 \\
\bottomrule
\end{tabular}
\caption{Case study 3 of answer stability and sensitivity. Purple marks the answer group used for A-STB; purple and green mark the semantic groups used for A-SST. \begin{CJK*}{UTF8}{gbsn}洛阳\end{CJK*} is the Chinese character of ``Luoyang''.}
\label{tab:case-stab-sens-detail-19}
\end{table*}

\subsection{Results of GPT-4o}

We report the aggregated results for GPT-4o in \cref{tab:gpt4o_results}. Due to limited access to the commercial model, we evaluate only five CE methods. Nevertheless, the observed trends are largely consistent with the main results reported in the paper. Note that, consistent with the main paper, we also use GPT-4o to judge the correctness of GPT-4o responses. However, since our proposed metrics do not depend on correctness labels, this setup does not affect our findings.

\begin{table*}[htb]
\centering
\setlength{\tabcolsep}{5pt}
\footnotesize

\begin{tabular}{lcccccc}
\toprule
Method & ECE $\downarrow$ & Brier $\downarrow$ & AUROC $\uparrow$ & P-RB $\uparrow$ & A-STB $\uparrow$ & A-SST $\uparrow$ \\
\midrule
Seq. Likelihood & 0.264 & 0.275 & 0.678 & 0.949 & 0.916 & 0.216 \\
Platt Scaling & 0.076 & 0.216 & 0.649 & 0.990 & 0.982 & 0.047 \\
Verbalized Conf. & 0.258 & 0.262 & 0.701 & 0.962 & 0.985 & 0.109 \\
P(True) & 0.295 & 0.296 & 0.733 & 0.976 & 0.988 & 0.226 \\
Calib1 & 0.104 & 0.208 & 0.704 & 0.982 & 0.960 & 0.091 \\
\bottomrule
\end{tabular}
\caption{GPT-4o results aggregated over all datasets. Due to the limited access of the commercial model, only methods that do not require information beyond the top 20 output logits apply to this model.}
\label{tab:gpt4o_results}
\end{table*}

\subsection{Robustness:  Confidence-elicit Prompts for Verbalized Confidence}\label[appendix]{sec:conf-prompt-vc}

In the main paper, we first use an \textit{answer-elicit prompt} (\textbf{Ans-elicit}) that instructs the LLM to produce an answer to the input question. At this stage, all the methods except verbalized confidence can already estimate confidence scores based on the generated answer. In contrast, verbalized confidence requires an additional \textit{confidence-elicit prompt} (\textbf{Conf-elicit}) to guide the LLM in verbalizing its confidence regarding the produced answer.
To further investigate the effect of prompt perturbations on confidence estimation, we conduct additional experiments for verbalized confidence that only perturb the confidence-elicit prompts instead.
This design allows us to explore the robustness of verbalized confidence under two prompt perturbation settings: (1) perturb the answer-elicit prompts and (2) perturb the confidence-elicit prompts.

\paragraph{Perturbation of confidence-elicitation prompt.} To  evaluate the robustness of verbalized confidence, we design seven variants of the confidence-elicitation prompt (see Table~\ref{tab:prompt_detail}), grouped into three types:
(i) Scale variants, which elicit confidence using numerical ranges of 0–1, 0–100\%, and 0–10;
(ii) Lexical variants, which vary the key term used to request a score between 0 and 1 (``probability'', ``certainty'' and ``confidence''); and
(iii) Linguistic expressions, following \citet{tian2023just}, which elicit confidence using predefined verbal scales from ``Almost No Chance'' to ``Almost certain'', and a corresponding multiple-choice format (e.g., ``a: Almost No Chance'').
Those selected expressions are converted to numerical scores according to the rules in the next paragraph.
All confidence outputs are normalized to a 0–1 scale for comparison. 

\begin{table*}
    \centering
    \footnotesize
    \renewcommand{\arraystretch}{1.3} 
    \begin{tabular}{p{1.6cm}cp{10cm}}
    \toprule
       \textbf{Concept}  &  \textbf{Prompt Name}& \multicolumn{1}{c}{\textbf{Prompt}}\\ 
         \hline
      Scale Variants    & P(1) & Provide the probability that your answer is correct. Give ONLY the probability \textbf{between 0.0 and 1.0}, no other words or explanation. \\
         & P(\%)  & Provide the probability that your answer is correct. Give ONLY the probability \textbf{between 0\% and 100\%}, no other words or explanation.\\
         & P(10) & Provide the probability that your answer is correct. Give ONLY the probability \textbf{between 0 and 10}, no other words or explanation.\\ 
         \hline
      Lexical Variants   & CF(1) & Provide the \textbf{confidence} that your answer is correct. Give ONLY the confidence {between 0.0 and 1.0}, no other words or explanation. \\
         & CT(1) & Provide the \textbf{certainty} that your answer is correct. Give ONLY the confidence {between 0 and 10}, no other words or explanation. \\
         \hline
      Linguistic Expressions   & L.  & Describe how likely it is that your answer is correct as \textbf{one of the following expressions}: ['Almost No Chance', 'Highly Unlikely', 'Chances are Slight', 'Little Chance', 'Unlikely', 'Probably Not', 'About Even', 'Better than Even', 'Likely', 'Probably', 'Very Good Chance','Highly Likely', 'Almost Certain']. Give ONLY the chosen expression, no other words or explanation. \\
         & L. MC  & Describe how likely it is that your answer is correct by choosing \textbf{one of the following options}: [a: 'Almost No Chance', b: 'Highly Unlikely', c: 'Chances are Slight', d: 'Little Chance', e:'Unlikely', f: 'Probably Not', g: 'About Even', h: 'Better than Even', i: 'Likely', j: 'Probably', k; 'Very Good Chance', l: 'Highly Likely', m: 'Almost Certain']. Give ONLY the chosen option, no other words or explanation.\\
    \bottomrule
    \end{tabular}
    \caption{Detail of prompts for eliciting confidence from LLMs. These prompts are specifically applied to verbalized confidence, we use these prompts to further validate the robustness of verbalized confidence in \cref{sec:conf-prompt-vc}.}
    \label{tab:prompt_detail}
\end{table*}

\begin{table}[hbp]
\centering
\begin{tabular}{lrr}
\toprule
Dataset & Ans-elicit  & Conf-elicit \\
\midrule
NQ & 0.985 & 0.907 \\
PopQA & 0.984 & 0.895 \\
SciQ & 0.990 & 0.928 \\
TriviaQA & 0.985 & 0.921 \\
AVG & 0.986 & 0.913 \\
\bottomrule
\end{tabular}
\caption{Robustness of verbalized confidence under two prompt perturbation settings: perturbing answer-elicit prompt (Ans-elicit) and perturbing confidence-elicit prompt (Conf-elicit).} \label{tab:vc_robustness_prompt_settings}
\end{table}
\paragraph{Mapping of Linguistic Prompts.} For the linguistic prompts, we followed \citet{tian2023just} to map the linguistic expressions to numerical scores. The specific mapping are [``Almost No Chance'', ``Highly Unlikely'', ``Chances are Slight'',
``Little Chance'', ``Unlikely'', ``Probably Not'', ``About Even'', ``Better than Even'',
``Likely'', ``Probably'', ``Very Good Chance'', ``Highly Likely'', 'Almost Certain''] to [0.02, 0.05, 0.1, 0.1, 0.2, 0.25, 0.5, 0.6, 0.7, 0.7, 0.8, 0.9, 0.95] according to \citet{fagen-ulmschneider2023perception}.

\paragraph{Results and Analysis.} As shown in Table~\ref{tab:vc_robustness_prompt_settings}, when perturb only the answer-elicit prompts (Ans-elicit), verbalized confidence produces more stable confidence estimates ($\text{P-RB}=0.986$) compared to perturb confidence-elicit prompts are used ($\text{P-RB}=0.913$). We hypothesize that this difference in stability arises because the generated answers remain largely consistent across perturbed answer-elicit prompts, leading to nearly identical instructions when subsequently eliciting confidence for those answers. In contrast, perturbations to the confidence-elicit prompts directly alter how the model is instructed to express its confidence, thereby introducing greater variability in the resulting confidence scores. 

\begin{figure*}[htb]
    \centering
     \begin{subfigure}[t]{0.95\linewidth}
        \centering
        \includegraphics[width=1\linewidth]{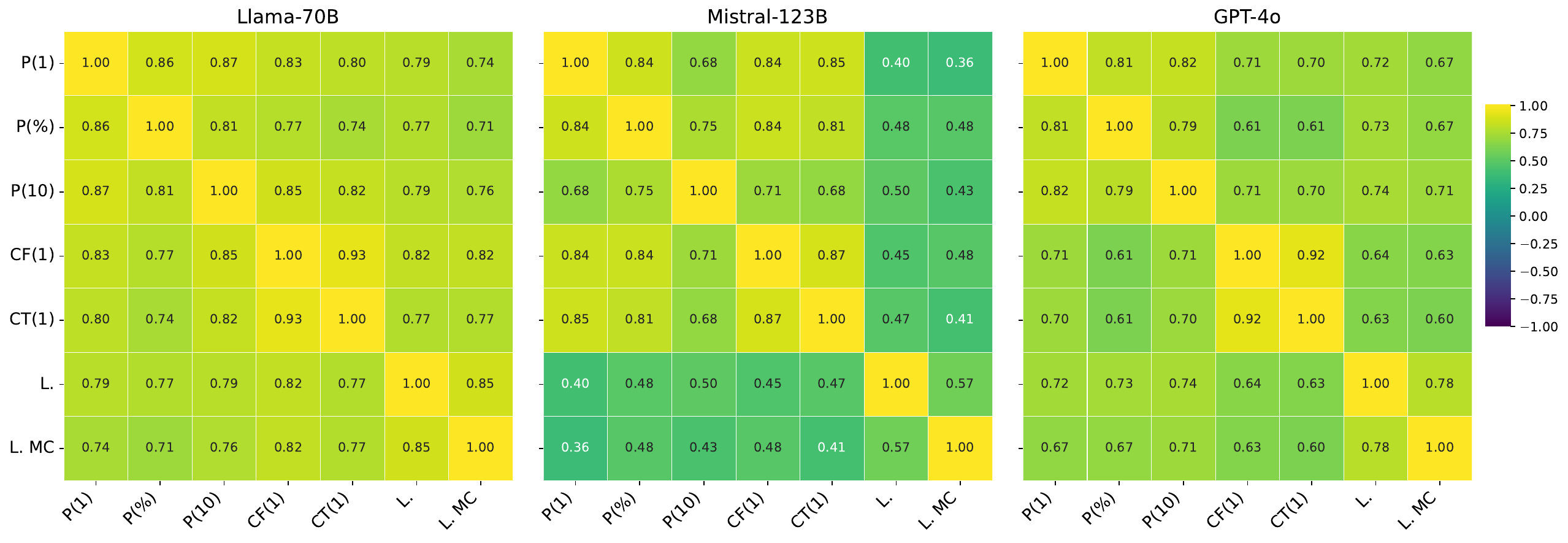}
        \end{subfigure}
        \begin{subfigure}[t]{0.95\linewidth}
        \centering
        \includegraphics[width=1\linewidth]{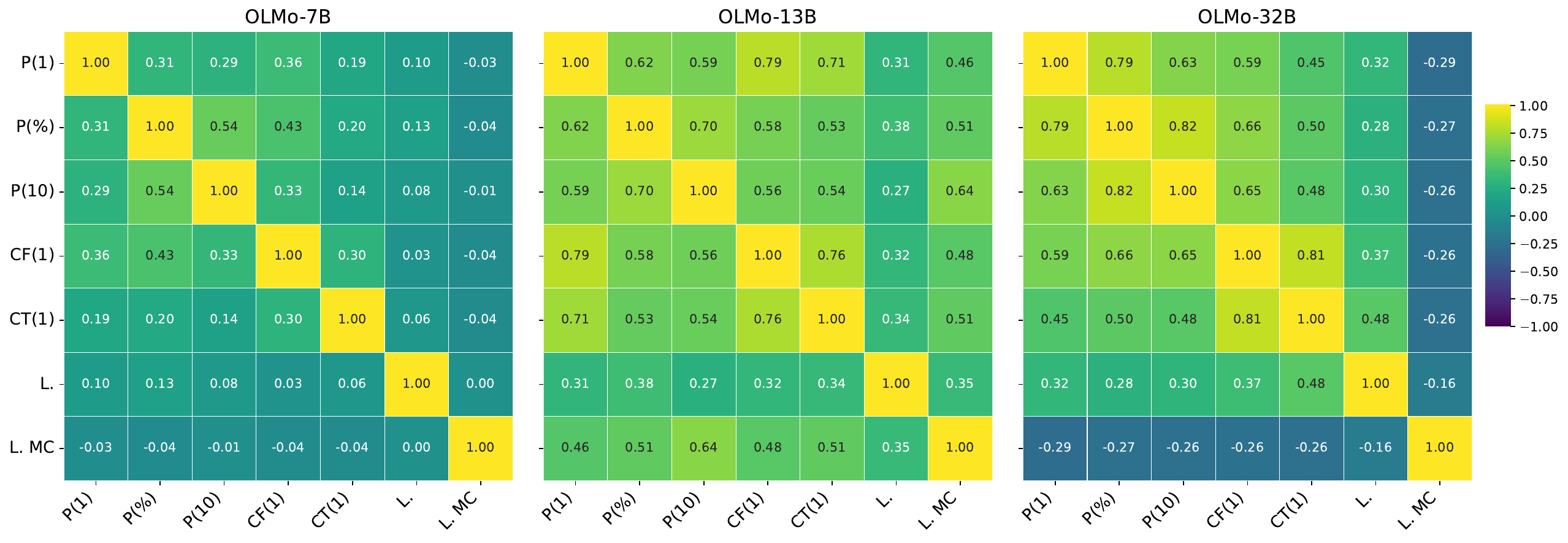}
        
        \end{subfigure}
        \caption{\textit{Robustness}: Pearson correlations of verbalized confidence when using different confidence-elicit prompts in \cref{tab:prompt_detail}.}
    \label{fig:corr_large_model}
\end{figure*}

\paragraph{Prompt-wise Confidence Analysis for Verbalized Confidence.} To further analyze the reason of the low robustness of verbalized confidence, we conduct a prompt-wise confidence analysis on selected models by examining the pairwise correlations of confidences between conf-elicit prompts for individual models (shown by  \cref{fig:corr_large_model}). We found that larger models maintain higher stability of confidence, demonstrating their ability to respond to diverse confidence-eliciting prompts.  LLMs remain stable with the change of scales and lexical words in prompts.  Specifically, prompts that perturbed with scale and lexical words change tend to have more correlated verbalized confidence compared to adding linguistic expressions in the prompt. Small models (<= 32B) struggle more in maintaining robustness, especially with the multi-choice linguistic expression prompts. 

\subsection{Sensitivity: Alternative Sets of Semantically Different Answers}\label{sec:sens_add}

\cref{tab:sensitivity_setting_comparison} reports sensitivity results obtained by using all semantically different answers outside $\hat{\mathcal{Y}}^{\text{max}}$ (denoted as $\hat{\mathcal{Y}}^{\notin \text{max}}$), instead of the original setting that compares $\hat{\mathcal{Y}}^{\text{max}}$ with $\hat{\mathcal{Y}}^{\text{min}}$. The results remain stable under this alternative definition and do not affect the main findings of the paper.

\begin{table}
\centering
\begin{tabular}{lrr}
\toprule
Method & $\hat{\mathcal{Y}}^{\text{min}}$ & $\hat{\mathcal{Y}}^{\notin \text{max}}$ \\
\midrule
Seq. Likelihood & 0.182 & 0.171 \\
Boosted Prob. & 0.037 & 0.035 \\
Platt Scaling & 0.039 & 0.037 \\
P(True) & 0.228 & 0.227 \\
Attention Score & 0.032 & 0.032 \\
Hidden Score & 0.034 & 0.034 \\
SAPLMA & 0.180 & 0.178 \\
P(IK) & 0.000 & 0.000 \\
Verbalized Conf. & 0.057 & 0.054 \\
Calib1 & 0.065 & 0.065 \\
\bottomrule
\end{tabular}
\caption{Comparison of answer sensitivity settings. The original setting compares the largest semantic group with the smallest semantic group $\hat{\mathcal{Y}}^{\text{min}}$; the alternative compares the largest group with all remaining semantically different answers $\hat{\mathcal{Y}}^{\notin \text{max}}$ .}
\label{tab:sensitivity_setting_comparison}
\end{table}

\subsection{Instruction vs Reasoning Models}\label{sec:reasoning-instuct}

We compare the averaged results of instruction-tuned and reasoning models across all datasets; the results are shown in \cref{tab:new_csv_dataset_metrics_nq}-\cref{tab:new_csv_dataset_metrics_popqa}. Overall, both model types exhibit highly similar trends. We hypothesize that this is because the evaluated tasks consist of short-form question answering, which requires limited multi-step reasoning and therefore yields only minor differences between instruction-tuned and reasoning models.

\begin{table*}
\centering\footnotesize
\sisetup{round-mode=places, round-precision=2, table-format=1.2, table-number-alignment=center}
\setlength{\tabcolsep}{5pt}
\renewcommand{\arraystretch}{1.1}
\resizebox{\linewidth}{!}{%
\begin{tabular}{l *{6}{S} @{\hspace{1em}} *{6}{S}@{}}
\toprule
& \multicolumn{6}{c}{\textbf{Instruction Model}} 
& \multicolumn{6}{c}{\textbf{Reasoning Model}}\\
\cmidrule(lr){2-7} \cmidrule(lr){8-13}

Method & {ECE} & {Brier} & {AUROC} & {P-RB} & {A-STB} & {A-SST}
       & {ECE} & {Brier} & {AUROC} & {P-RB} & {A-STB} & {A-SST} \\
\midrule
Seq. Likelihood & \cellcolor[HTML]{B3CBE3}0.410 & \cellcolor[HTML]{85AAD0}0.387 & \cellcolor[HTML]{3B76B1}0.707 & \cellcolor[HTML]{5487BB}0.952 & \cellcolor[HTML]{C5D7EA}0.975 & \cellcolor[HTML]{2C6BAB}0.169 & \cellcolor[HTML]{8EB0D3}0.378 & \cellcolor[HTML]{5286BB}0.321 & \cellcolor[HTML]{2A69AA}0.750 & \cellcolor[HTML]{6593C2}0.949 & \cellcolor[HTML]{D5E3F1}0.938 & \cellcolor[HTML]{4D82B9}0.146 \\
Boosted Prob. & \cellcolor[HTML]{F7FBFF}0.556 & \cellcolor[HTML]{F7FBFF}0.536 & \cellcolor[HTML]{3470AE}0.712 & \cellcolor[HTML]{09529D}0.989 & \cellcolor[HTML]{2A69AA}0.996 & \cellcolor[HTML]{D1E0EF}0.031 & \cellcolor[HTML]{ECF3FA}0.615 & \cellcolor[HTML]{DCE8F4}0.568 & \cellcolor[HTML]{5A8BBE}0.715 & \cellcolor[HTML]{1A5EA3}0.982 & \cellcolor[HTML]{5689BC}0.976 & \cellcolor[HTML]{C7D9EB}0.041 \\
Platt Scaling & \cellcolor[HTML]{588ABD}0.217 & \cellcolor[HTML]{306DAC}0.275 & \cellcolor[HTML]{3B76B1}0.707 & \cellcolor[HTML]{08519C}0.990 & \cellcolor[HTML]{2F6CAC}0.995 & \cellcolor[HTML]{CDDDED}0.035 & \cellcolor[HTML]{6392C2}0.270 & \cellcolor[HTML]{3470AE}0.266 & \cellcolor[HTML]{2A69AA}0.750 & \cellcolor[HTML]{08519C}0.990 & \cellcolor[HTML]{326FAD}0.987 & \cellcolor[HTML]{D4E2F0}0.030 \\
P(True) & \cellcolor[HTML]{8EB0D4}0.332 & \cellcolor[HTML]{6996C4}0.350 & \cellcolor[HTML]{2E6CAC}0.716 & \cellcolor[HTML]{8AADD2}0.925 & \cellcolor[HTML]{EEF5FB}0.970 & \cellcolor[HTML]{1359A1}0.190 & \cellcolor[HTML]{97B6D7}0.401 & \cellcolor[HTML]{6B98C5}0.366 & \cellcolor[HTML]{3470AE}0.742 & \cellcolor[HTML]{6A97C5}0.947 & \cellcolor[HTML]{CEDEEE}0.940 & \cellcolor[HTML]{08519C}0.205 \\
Attention Score & \cellcolor[HTML]{08519C}0.045 & \cellcolor[HTML]{10579F}0.233 & \cellcolor[HTML]{F7FBFF}0.575 & \cellcolor[HTML]{0E559E}0.987 & \cellcolor[HTML]{3470AE}0.994 & \cellcolor[HTML]{D7E4F2}0.027 & \cellcolor[HTML]{08519C}0.042 & \cellcolor[HTML]{0E559F}0.199 & \cellcolor[HTML]{E8F0F9}0.614 & \cellcolor[HTML]{1158A0}0.986 & \cellcolor[HTML]{3571AE}0.987 & \cellcolor[HTML]{D5E3F1}0.030 \\
Hidden Score & \cellcolor[HTML]{0A539D}0.049 & \cellcolor[HTML]{0C549E}0.228 & \cellcolor[HTML]{C3D6EA}0.612 & \cellcolor[HTML]{195DA3}0.981 & \cellcolor[HTML]{447BB5}0.992 & \cellcolor[HTML]{C1D4E9}0.045 & \cellcolor[HTML]{09529C}0.044 & \cellcolor[HTML]{0F569F}0.200 & \cellcolor[HTML]{F7FBFF}0.603 & \cellcolor[HTML]{0F569F}0.987 & \cellcolor[HTML]{3B75B1}0.985 & \cellcolor[HTML]{DDE8F4}0.022 \\
SAPLMA & \cellcolor[HTML]{6E9AC6}0.263 & \cellcolor[HTML]{316EAD}0.277 & \cellcolor[HTML]{08519C}0.743 & \cellcolor[HTML]{F7FBFF}0.870 & \cellcolor[HTML]{F7FBFF}0.969 & \cellcolor[HTML]{08519C}0.199 & \cellcolor[HTML]{5185BA}0.226 & \cellcolor[HTML]{2264A7}0.235 & \cellcolor[HTML]{08519C}0.774 & \cellcolor[HTML]{F7FBFF}0.885 & \cellcolor[HTML]{F7FBFF}0.928 & \cellcolor[HTML]{4F83B9}0.145 \\
P(IK) & \cellcolor[HTML]{5487BB}0.207 & \cellcolor[HTML]{2F6DAC}0.274 & \cellcolor[HTML]{A8C3DE}0.631 & \cellcolor[HTML]{487FB7}0.958 & \cellcolor[HTML]{08519C}1.000 & \cellcolor[HTML]{F7FBFF}0.000 & \cellcolor[HTML]{4079B3}0.182 & \cellcolor[HTML]{2364A7}0.236 & \cellcolor[HTML]{7FA6CD}0.689 & \cellcolor[HTML]{477EB6}0.962 & \cellcolor[HTML]{08519C}1.000 & \cellcolor[HTML]{F7FBFF}0.000 \\
Verbalized Conf. & \cellcolor[HTML]{E8F1F9}0.524 & \cellcolor[HTML]{E4EDF7}0.511 & \cellcolor[HTML]{D9E5F2}0.597 & \cellcolor[HTML]{1359A0}0.985 & \cellcolor[HTML]{306DAC}0.995 & \cellcolor[HTML]{B5CCE4}0.055 & \cellcolor[HTML]{F7FBFF}0.644 & \cellcolor[HTML]{F7FBFF}0.617 & \cellcolor[HTML]{CFDEEE}0.632 & \cellcolor[HTML]{0D559E}0.987 & \cellcolor[HTML]{437BB4}0.982 & \cellcolor[HTML]{D8E5F2}0.027 \\
Calib1 & \cellcolor[HTML]{1359A0}0.067 & \cellcolor[HTML]{08519C}0.223 & \cellcolor[HTML]{95B6D7}0.644 & \cellcolor[HTML]{155AA1}0.984 & \cellcolor[HTML]{4E83B9}0.991 & \cellcolor[HTML]{A3BFDC}0.070 & \cellcolor[HTML]{0C549E}0.051 & \cellcolor[HTML]{08519C}0.187 & \cellcolor[HTML]{5E8EC0}0.713 & \cellcolor[HTML]{1D60A5}0.981 & \cellcolor[HTML]{85AAD0}0.962 & \cellcolor[HTML]{BBD0E6}0.051 \\
\bottomrule
\end{tabular}}
\caption{NQ metric values averaged across instruction and reasoning models, respectively. Darker color indicates better performance.}
\label{tab:new_csv_dataset_metrics_nq}
\end{table*}

\begin{table*}
\centering\footnotesize
\sisetup{round-mode=places, round-precision=2, table-format=1.2, table-number-alignment=center}
\setlength{\tabcolsep}{5pt}
\renewcommand{\arraystretch}{1.1}
\resizebox{\linewidth}{!}{%
\begin{tabular}{l *{6}{S} @{\hspace{1em}} *{6}{S}@{}}
\toprule
& \multicolumn{6}{c}{\textbf{Instruction Model}} 
& \multicolumn{6}{c}{\textbf{Reasoning Model}}\\
\cmidrule(lr){2-7} \cmidrule(lr){8-13}

Method & {ECE} & {Brier} & {AUROC} & {P-RB} & {A-STB} & {A-SST}
       & {ECE} & {Brier} & {AUROC} & {P-RB} & {A-STB} & {A-SST} \\

\midrule
Seq. Likelihood & \cellcolor[HTML]{729CC8}0.087 & \cellcolor[HTML]{326FAD}0.149 & \cellcolor[HTML]{6190C1}0.677 & \cellcolor[HTML]{A3BFDC}0.937 & \cellcolor[HTML]{F7FBFF}0.969 & \cellcolor[HTML]{3974B0}0.172 & \cellcolor[HTML]{2767A9}0.038 & \cellcolor[HTML]{08519C}0.151 & \cellcolor[HTML]{3571AF}0.705 & \cellcolor[HTML]{598ABD}0.942 & \cellcolor[HTML]{F7FBFF}0.941 & \cellcolor[HTML]{3E77B2}0.147 \\
Boosted Prob. & \cellcolor[HTML]{F7FBFF}0.177 & \cellcolor[HTML]{F7FBFF}0.179 & \cellcolor[HTML]{2A69AA}0.723 & \cellcolor[HTML]{08519C}0.995 & \cellcolor[HTML]{0E559E}0.999 & \cellcolor[HTML]{EDF4FB}0.010 & \cellcolor[HTML]{F0F6FC}0.175 & \cellcolor[HTML]{D4E2F1}0.184 & \cellcolor[HTML]{1258A0}0.732 & \cellcolor[HTML]{08519C}0.992 & \cellcolor[HTML]{155BA2}0.996 & \cellcolor[HTML]{E2ECF6}0.017 \\
Platt Scaling & \cellcolor[HTML]{81A7CE}0.097 & \cellcolor[HTML]{5286BB}0.154 & \cellcolor[HTML]{6190C1}0.677 & \cellcolor[HTML]{1E61A5}0.986 & \cellcolor[HTML]{3E78B2}0.993 & \cellcolor[HTML]{CBDCED}0.040 & \cellcolor[HTML]{769FCA}0.092 & \cellcolor[HTML]{477EB6}0.161 & \cellcolor[HTML]{3571AF}0.705 & \cellcolor[HTML]{10579F}0.987 & \cellcolor[HTML]{3E77B2}0.986 & \cellcolor[HTML]{CBDCED}0.035 \\
P(True) & \cellcolor[HTML]{E3ECF7}0.163 & \cellcolor[HTML]{BED2E7}0.170 & \cellcolor[HTML]{2767A9}0.726 & \cellcolor[HTML]{83A8CF}0.949 & \cellcolor[HTML]{457DB5}0.992 & \cellcolor[HTML]{2F6DAC}0.182 & \cellcolor[HTML]{91B3D5}0.111 & \cellcolor[HTML]{1359A1}0.153 & \cellcolor[HTML]{08519C}0.739 & \cellcolor[HTML]{2A69AA}0.971 & \cellcolor[HTML]{3370AE}0.989 & \cellcolor[HTML]{08519C}0.190 \\
Attention Score & \cellcolor[HTML]{08519C}0.017 & \cellcolor[HTML]{3470AE}0.150 & \cellcolor[HTML]{F7FBFF}0.551 & \cellcolor[HTML]{1158A0}0.991 & \cellcolor[HTML]{1B5FA4}0.997 & \cellcolor[HTML]{E2ECF6}0.019 & \cellcolor[HTML]{08519C}0.017 & \cellcolor[HTML]{3F78B3}0.160 & \cellcolor[HTML]{F7FBFF}0.560 & \cellcolor[HTML]{0A529D}0.991 & \cellcolor[HTML]{195DA3}0.996 & \cellcolor[HTML]{E4EDF7}0.015 \\
Hidden Score & \cellcolor[HTML]{2867A9}0.038 & \cellcolor[HTML]{2B6AAA}0.148 & \cellcolor[HTML]{D0DFEF}0.584 & \cellcolor[HTML]{2E6CAC}0.981 & \cellcolor[HTML]{2868A9}0.996 & \cellcolor[HTML]{C6D8EB}0.044 & \cellcolor[HTML]{2162A6}0.034 & \cellcolor[HTML]{2A69AA}0.156 & \cellcolor[HTML]{B7CEE5}0.607 & \cellcolor[HTML]{2163A6}0.977 & \cellcolor[HTML]{3E77B2}0.987 & \cellcolor[HTML]{C1D5E9}0.043 \\
SAPLMA & \cellcolor[HTML]{CCDCED}0.148 & \cellcolor[HTML]{9AB9D9}0.165 & \cellcolor[HTML]{08519C}0.752 & \cellcolor[HTML]{F7FBFF}0.906 & \cellcolor[HTML]{85AAD0}0.984 & \cellcolor[HTML]{08519C}0.217 & \cellcolor[HTML]{F7FBFF}0.180 & \cellcolor[HTML]{F7FBFF}0.189 & \cellcolor[HTML]{2465A7}0.719 & \cellcolor[HTML]{F7FBFF}0.842 & \cellcolor[HTML]{D2E0EF}0.950 & \cellcolor[HTML]{145AA1}0.180 \\
P(IK) & \cellcolor[HTML]{3B75B1}0.051 & \cellcolor[HTML]{08519C}0.143 & \cellcolor[HTML]{5487BB}0.688 & \cellcolor[HTML]{3A74B1}0.976 & \cellcolor[HTML]{08519C}1.000 & \cellcolor[HTML]{F7FBFF}0.000 & \cellcolor[HTML]{8DB0D3}0.108 & \cellcolor[HTML]{6391C1}0.165 & \cellcolor[HTML]{6A97C5}0.666 & \cellcolor[HTML]{2163A6}0.976 & \cellcolor[HTML]{08519C}1.000 & \cellcolor[HTML]{F7FBFF}0.000 \\
Verbalized Conf. & \cellcolor[HTML]{C4D7EA}0.143 & \cellcolor[HTML]{9EBCDA}0.166 & \cellcolor[HTML]{A7C2DE}0.618 & \cellcolor[HTML]{165BA2}0.990 & \cellcolor[HTML]{1157A0}0.999 & \cellcolor[HTML]{D2E0F0}0.034 & \cellcolor[HTML]{DDE8F4}0.162 & \cellcolor[HTML]{B1C9E2}0.178 & \cellcolor[HTML]{2F6DAC}0.710 & \cellcolor[HTML]{09519C}0.992 & \cellcolor[HTML]{185CA3}0.996 & \cellcolor[HTML]{D5E3F1}0.027 \\
Calib1 & \cellcolor[HTML]{749EC9}0.089 & \cellcolor[HTML]{6593C3}0.157 & \cellcolor[HTML]{BFD3E8}0.598 & \cellcolor[HTML]{2666A8}0.984 & \cellcolor[HTML]{316EAD}0.995 & \cellcolor[HTML]{A7C2DE}0.073 & \cellcolor[HTML]{9DBBDA}0.118 & \cellcolor[HTML]{80A7CE}0.170 & \cellcolor[HTML]{AEC7E1}0.614 & \cellcolor[HTML]{0A539D}0.991 & \cellcolor[HTML]{4E83B9}0.982 & \cellcolor[HTML]{C0D4E8}0.043 \\
\bottomrule
\end{tabular}}
\caption{SciQ metric values averaged across instruction and reasoning models, respectively. Darker color indicates better performance.}
\label{tab:new_csv_dataset_metrics_sciq}
\end{table*}

\begin{table*}
\centering\footnotesize
\sisetup{round-mode=places, round-precision=2, table-format=1.2, table-number-alignment=center}
\setlength{\tabcolsep}{5pt}
\renewcommand{\arraystretch}{1.1}
\resizebox{\linewidth}{!}{%
\begin{tabular}{l *{6}{S} @{\hspace{1em}} *{6}{S}@{}}
\toprule
& \multicolumn{6}{c}{\textbf{Instruction Model}} 
& \multicolumn{6}{c}{\textbf{Reasoning Model}}\\
\cmidrule(lr){2-7} \cmidrule(lr){8-13}

Method & {ECE} & {Brier} & {AUROC} & {P-RB} & {A-STB} & {A-SST}
       & {ECE} & {Brier} & {AUROC} & {P-RB} & {A-STB} & {A-SST} \\
\midrule
Seq. Likelihood & \cellcolor[HTML]{B8CEE5}0.277 & \cellcolor[HTML]{6E9AC6}0.285 & \cellcolor[HTML]{185DA3}0.688 & \cellcolor[HTML]{5A8BBE}0.951 & \cellcolor[HTML]{F7FBFF}0.985 & \cellcolor[HTML]{497FB7}0.206 & \cellcolor[HTML]{8DB0D3}0.266 & \cellcolor[HTML]{2C6BAB}0.272 & \cellcolor[HTML]{0C549E}0.757 & \cellcolor[HTML]{4B81B8}0.945 & \cellcolor[HTML]{94B5D6}0.960 & \cellcolor[HTML]{5789BD}0.173 \\
Boosted Prob. & \cellcolor[HTML]{F7FBFF}0.370 & \cellcolor[HTML]{F7FBFF}0.361 & \cellcolor[HTML]{08519C}0.701 & \cellcolor[HTML]{0C549E}0.993 & \cellcolor[HTML]{195DA3}0.999 & \cellcolor[HTML]{DFEAF5}0.029 & \cellcolor[HTML]{EAF2FA}0.434 & \cellcolor[HTML]{CADBEC}0.410 & \cellcolor[HTML]{08519C}0.761 & \cellcolor[HTML]{0D559E}0.986 & \cellcolor[HTML]{1C5FA4}0.994 & \cellcolor[HTML]{CADBEC}0.049 \\
Platt Scaling & \cellcolor[HTML]{316EAD}0.079 & \cellcolor[HTML]{0F569F}0.232 & \cellcolor[HTML]{185DA3}0.688 & \cellcolor[HTML]{1258A0}0.990 & \cellcolor[HTML]{3974B0}0.997 & \cellcolor[HTML]{D0DFEF}0.046 & \cellcolor[HTML]{5C8DBF}0.177 & \cellcolor[HTML]{165BA2}0.252 & \cellcolor[HTML]{0C549E}0.757 & \cellcolor[HTML]{0A539D}0.988 & \cellcolor[HTML]{2566A8}0.991 & \cellcolor[HTML]{D3E1F0}0.039 \\
P(True) & \cellcolor[HTML]{AEC7E1}0.263 & \cellcolor[HTML]{6492C2}0.279 & \cellcolor[HTML]{1B5EA4}0.686 & \cellcolor[HTML]{6795C4}0.944 & \cellcolor[HTML]{80A6CE}0.992 & \cellcolor[HTML]{08519C}0.283 & \cellcolor[HTML]{99B8D8}0.288 & \cellcolor[HTML]{497FB7}0.297 & \cellcolor[HTML]{1057A0}0.753 & \cellcolor[HTML]{3571AF}0.960 & \cellcolor[HTML]{3672AF}0.986 & \cellcolor[HTML]{08519C}0.259 \\
Attention Score & \cellcolor[HTML]{10579F}0.031 & \cellcolor[HTML]{0C549E}0.230 & \cellcolor[HTML]{BAD0E6}0.564 & \cellcolor[HTML]{145AA1}0.989 & \cellcolor[HTML]{2565A8}0.998 & \cellcolor[HTML]{D9E6F3}0.035 & \cellcolor[HTML]{10579F}0.040 & \cellcolor[HTML]{1258A0}0.249 & \cellcolor[HTML]{F7FBFF}0.535 & \cellcolor[HTML]{0C549E}0.987 & \cellcolor[HTML]{1C5FA4}0.994 & \cellcolor[HTML]{DEE9F5}0.027 \\
Hidden Score & \cellcolor[HTML]{08519C}0.019 & \cellcolor[HTML]{10569F}0.232 & \cellcolor[HTML]{F7FBFF}0.517 & \cellcolor[HTML]{08519C}0.995 & \cellcolor[HTML]{0D549E}0.999 & \cellcolor[HTML]{EBF3FA}0.014 & \cellcolor[HTML]{08519C}0.025 & \cellcolor[HTML]{1157A0}0.247 & \cellcolor[HTML]{F3F8FD}0.538 & \cellcolor[HTML]{08519C}0.990 & \cellcolor[HTML]{1A5DA3}0.995 & \cellcolor[HTML]{E1ECF6}0.024 \\
SAPLMA & \cellcolor[HTML]{A8C3DE}0.255 & \cellcolor[HTML]{6D99C6}0.284 & \cellcolor[HTML]{3E77B2}0.659 & \cellcolor[HTML]{F7FBFF}0.867 & \cellcolor[HTML]{F2F8FD}0.985 & \cellcolor[HTML]{2D6BAB}0.239 & \cellcolor[HTML]{86ABD0}0.253 & \cellcolor[HTML]{3974B0}0.282 & \cellcolor[HTML]{3B75B1}0.713 & \cellcolor[HTML]{F7FBFF}0.830 & \cellcolor[HTML]{F7FBFF}0.932 & \cellcolor[HTML]{5386BB}0.178 \\
P(IK) & \cellcolor[HTML]{3B75B1}0.094 & \cellcolor[HTML]{08519C}0.228 & \cellcolor[HTML]{6795C3}0.628 & \cellcolor[HTML]{4B80B8}0.959 & \cellcolor[HTML]{08519C}1.000 & \cellcolor[HTML]{F7FBFF}0.000 & \cellcolor[HTML]{487EB6}0.141 & \cellcolor[HTML]{08519C}0.240 & \cellcolor[HTML]{427AB4}0.706 & \cellcolor[HTML]{4179B4}0.952 & \cellcolor[HTML]{08519C}1.000 & \cellcolor[HTML]{F7FBFF}0.000 \\
Verbalized Conf. & \cellcolor[HTML]{E8F0F9}0.348 & \cellcolor[HTML]{E2ECF6}0.350 & \cellcolor[HTML]{99B8D8}0.589 & \cellcolor[HTML]{1C5FA4}0.984 & \cellcolor[HTML]{1E60A5}0.998 & \cellcolor[HTML]{A9C4DF}0.092 & \cellcolor[HTML]{F7FBFF}0.458 & \cellcolor[HTML]{F7FBFF}0.450 & \cellcolor[HTML]{729DC8}0.660 & \cellcolor[HTML]{0A529D}0.989 & \cellcolor[HTML]{165BA2}0.996 & \cellcolor[HTML]{CFDFEF}0.043 \\
Calib1 & \cellcolor[HTML]{2565A8}0.062 & \cellcolor[HTML]{0C549D}0.230 & \cellcolor[HTML]{9EBBDA}0.586 & \cellcolor[HTML]{10579F}0.991 & \cellcolor[HTML]{2465A8}0.998 & \cellcolor[HTML]{CBDBED}0.053 & \cellcolor[HTML]{2566A8}0.078 & \cellcolor[HTML]{0B539D}0.242 & \cellcolor[HTML]{91B2D5}0.631 & \cellcolor[HTML]{1157A0}0.984 & \cellcolor[HTML]{3F78B3}0.984 & \cellcolor[HTML]{BDD1E7}0.063 \\
\bottomrule
\end{tabular}}
\caption{TriviaQA metric values averaged across instruction and reasoning models, respectively. Darker color indicates better performance.}
\label{tab:new_csv_dataset_metrics_triviaqa}
\end{table*}

\begin{table*}
\centering\footnotesize
\sisetup{round-mode=places, round-precision=2, table-format=1.2, table-number-alignment=center}
\setlength{\tabcolsep}{5pt}
\renewcommand{\arraystretch}{1.1}
\resizebox{\linewidth}{!}{%
\begin{tabular}{l *{6}{S} @{\hspace{1em}} *{6}{S}@{}}
\toprule
& \multicolumn{6}{c}{\textbf{Instruction Model}} 
& \multicolumn{6}{c}{\textbf{Reasoning Model}}\\
\cmidrule(lr){2-7} \cmidrule(lr){8-13}

Method & {ECE} & {Brier} & {AUROC} & {P-RB} & {A-STB} & {A-SST}
       & {ECE} & {Brier} & {AUROC} & {P-RB} & {A-STB} & {A-SST} \\
\midrule
Seq. Likelihood & \cellcolor[HTML]{A8C3DE}0.407 & \cellcolor[HTML]{739DC8}0.334 & \cellcolor[HTML]{2B6AAB}0.832 & \cellcolor[HTML]{94B4D6}0.950 & \cellcolor[HTML]{F7FBFF}0.988 & \cellcolor[HTML]{3873B0}0.206 & \cellcolor[HTML]{6694C3}0.301 & \cellcolor[HTML]{2B6AAB}0.209 & \cellcolor[HTML]{0C549E}0.877 & \cellcolor[HTML]{A1BEDB}0.955 & \cellcolor[HTML]{F7FBFF}0.971 & \cellcolor[HTML]{5487BB}0.157 \\
Boosted Prob. & \cellcolor[HTML]{F7FBFF}0.584 & \cellcolor[HTML]{E5EEF7}0.531 & \cellcolor[HTML]{175CA2}0.857 & \cellcolor[HTML]{2D6BAB}0.982 & \cellcolor[HTML]{5185BA}0.996 & \cellcolor[HTML]{B8CEE5}0.068 & \cellcolor[HTML]{BAD0E6}0.537 & \cellcolor[HTML]{8AADD2}0.427 & \cellcolor[HTML]{08519C}0.882 & \cellcolor[HTML]{6E9AC6}0.968 & \cellcolor[HTML]{B1C9E2}0.980 & \cellcolor[HTML]{98B7D8}0.091 \\
Platt Scaling & \cellcolor[HTML]{6895C4}0.264 & \cellcolor[HTML]{457DB5}0.257 & \cellcolor[HTML]{2B6AAB}0.832 & \cellcolor[HTML]{1258A0}0.990 & \cellcolor[HTML]{3672AF}0.998 & \cellcolor[HTML]{D1E0EF}0.041 & \cellcolor[HTML]{6F9AC7}0.328 & \cellcolor[HTML]{3772AF}0.236 & \cellcolor[HTML]{0C549E}0.877 & \cellcolor[HTML]{1459A1}0.991 & \cellcolor[HTML]{3873B0}0.994 & \cellcolor[HTML]{D5E3F1}0.033 \\
P(True) & \cellcolor[HTML]{7BA3CC}0.307 & \cellcolor[HTML]{6291C1}0.307 & \cellcolor[HTML]{467DB6}0.797 & \cellcolor[HTML]{F2F8FD}0.921 & \cellcolor[HTML]{D3E1F0}0.990 & \cellcolor[HTML]{08519C}0.257 & \cellcolor[HTML]{91B2D5}0.421 & \cellcolor[HTML]{6A97C5}0.354 & \cellcolor[HTML]{477EB6}0.795 & \cellcolor[HTML]{ABC5E0}0.953 & \cellcolor[HTML]{ABC5DF}0.980 & \cellcolor[HTML]{08519C}0.229 \\
Attention Score & \cellcolor[HTML]{0A539D}0.053 & \cellcolor[HTML]{2767A9}0.205 & \cellcolor[HTML]{D8E5F2}0.612 & \cellcolor[HTML]{2062A6}0.986 & \cellcolor[HTML]{447CB5}0.997 & \cellcolor[HTML]{CADBEC}0.049 & \cellcolor[HTML]{0A529D}0.043 & \cellcolor[HTML]{1C5FA4}0.174 & \cellcolor[HTML]{C4D7EA}0.623 & \cellcolor[HTML]{1A5EA4}0.989 & \cellcolor[HTML]{467DB6}0.992 & \cellcolor[HTML]{D5E3F1}0.033 \\
Hidden Score & \cellcolor[HTML]{08519C}0.049 & \cellcolor[HTML]{2A69AA}0.210 & \cellcolor[HTML]{F7FBFF}0.573 & \cellcolor[HTML]{08519C}0.993 & \cellcolor[HTML]{3471AE}0.998 & \cellcolor[HTML]{D6E4F1}0.035 & \cellcolor[HTML]{08519C}0.038 & \cellcolor[HTML]{1D60A5}0.178 & \cellcolor[HTML]{F7FBFF}0.552 & \cellcolor[HTML]{08519C}0.994 & \cellcolor[HTML]{306EAD}0.995 & \cellcolor[HTML]{E1EBF6}0.021 \\
SAPLMA & \cellcolor[HTML]{2E6CAC}0.133 & \cellcolor[HTML]{08519C}0.152 & \cellcolor[HTML]{08519C}0.876 & \cellcolor[HTML]{F7FBFF}0.920 & \cellcolor[HTML]{A5C1DD}0.992 & \cellcolor[HTML]{8EB0D3}0.113 & \cellcolor[HTML]{316EAD}0.152 & \cellcolor[HTML]{1359A0}0.153 & \cellcolor[HTML]{0C549E}0.875 & \cellcolor[HTML]{F7FBFF}0.934 & \cellcolor[HTML]{F4F9FE}0.972 & \cellcolor[HTML]{B6CDE4}0.063 \\
P(IK) & \cellcolor[HTML]{165BA2}0.080 & \cellcolor[HTML]{0E559F}0.162 & \cellcolor[HTML]{3873B0}0.816 & \cellcolor[HTML]{6492C2}0.965 & \cellcolor[HTML]{08519C}1.000 & \cellcolor[HTML]{F7FBFF}0.000 & \cellcolor[HTML]{185CA3}0.083 & \cellcolor[HTML]{08519C}0.129 & \cellcolor[HTML]{1A5EA3}0.857 & \cellcolor[HTML]{5084BA}0.976 & \cellcolor[HTML]{08519C}1.000 & \cellcolor[HTML]{F7FBFF}0.000 \\
Verbalized Conf. & \cellcolor[HTML]{F7FBFF}0.585 & \cellcolor[HTML]{F7FBFF}0.562 & \cellcolor[HTML]{BAD0E6}0.650 & \cellcolor[HTML]{2868A9}0.983 & \cellcolor[HTML]{4E82B9}0.996 & \cellcolor[HTML]{AFC8E1}0.077 & \cellcolor[HTML]{F7FBFF}0.708 & \cellcolor[HTML]{F7FBFF}0.677 & \cellcolor[HTML]{97B7D7}0.684 & \cellcolor[HTML]{1F62A6}0.988 & \cellcolor[HTML]{2566A8}0.996 & \cellcolor[HTML]{DAE6F3}0.028 \\
Calib1 & \cellcolor[HTML]{175CA2}0.083 & \cellcolor[HTML]{165BA2}0.175 & \cellcolor[HTML]{5D8EBF}0.768 & \cellcolor[HTML]{2565A8}0.984 & \cellcolor[HTML]{437BB5}0.997 & \cellcolor[HTML]{B2CAE2}0.075 & \cellcolor[HTML]{2062A6}0.106 & \cellcolor[HTML]{0B539D}0.136 & \cellcolor[HTML]{3772AF}0.817 & \cellcolor[HTML]{1A5DA3}0.989 & \cellcolor[HTML]{86ABD0}0.985 & \cellcolor[HTML]{B4CBE3}0.064 \\
\bottomrule
\end{tabular}}
\caption{PopQA metric values averaged across instruction and reasoning models, respectively. Darker color indicates better performance.}
\label{tab:new_csv_dataset_metrics_popqa}
\end{table*}

\newpage

\section{Reproducibility Details}
\subsection{Use of AI Assistants}

AI-based assistants were used to support code debugging, language polishing, grammar correction, and improving the clarity of presentation. The authors developed and verified all scientific contributions, experimental design, analyses, and conclusions. AI assistants were not used to generate experimental results or annotations without human oversight and verification.

\subsection{Technical Details}
We use the vLLM \cite{vllm2023} library for LLM inference and serving.  All our experiments are conducted on NVIDIA HGX H100, which requires approximately 600 GPU hours to replicate.

\subsection{Detailed Answer-elicit Prompts}\label[appendix]{sec:ans-prompt}
\begin{table*}[htb]
    \centering
    \footnotesize
    \renewcommand{\arraystretch}{1.3} 
    \begin{tabular}{p{1.5cm}cp{10cm}}
    \toprule
      \textbf{Concept}  &  \textbf{Prompt Name}& \multicolumn{1}{c}{\textbf{Prompt}}\\ 
         \hline
       \multirow{5}{*}{Answer} & A1 & Answer the question, give ONLY the answer, no other words or explanation:\\
         & A2  & Answer the question, give ONLY the answer without explanation:\\
         & A3 & Answer the question, give ONLY the answer:\\ 
   
         & A4 & Answer the question as short as possible: \\
         & A5 & Answer the question with minimal words: \\
         \hline
     \multirow{6}{*}{Provide} & P1  & Provide an answer for the question, give ONLY the answer, no other words or explanation:\\
         & P2  & Provide an answer for the question without explanation:\\
   
     & P3 & Provide an answer for the question, give ONLY the answer: \\
         & P4 & Provide as short an answer as possible to the question:\\
         & P5 & Provide an answer with minimal words for the question:\\
    \bottomrule
    \end{tabular}
    \caption{Detail of prompts for eliciting answers from LLMs. These are the prompts used for evaluating robustness in the main paper.}
    \label{tab:ans_prompt_detail}
\end{table*}

In the main paper, all the CE methods are evaluated under the prompt variations in \cref{tab:ans_prompt_detail} for robustness.

\subsection{LLM-as-a-Judge Prompts} \label[appendix]{sec:prompt}
The prompts used for GPT-4o as a judge model are shown in \cref{prompt:correctness} and \cref{prompt:grouping}.

\begin{table*}[hbp]
\centering
\begin{tcolorbox}[
    width=\textwidth,
    colback=white,
    colframe=black,
    title=Semantic Equivalence Judgment Prompt for Correctness,
    fonttitle=\bfseries,
    boxrule=0.5pt,
    arc=2pt,
    left=4pt,
    right=4pt,
    top=4pt,
    bottom=4pt,
]
\small\ttfamily
You are evaluating whether an answer is semantically equivalent to the reference answer.
Judge meaning rather than exact wording. Two answers should be considered equivalent
if they express the same meaning, even when using different phrasing, synonyms, or
slightly different names referring to the same entity.\\

Also, treat answers as equivalent even if they include expressions of confidence or uncertainty (e.g., 'I'm pretty sure it's X' vs. 'X'). Focus on the core meaning of the answers and ignore differences in:\\
- wording or phrasing\\
- length\\
- fillers\\
- strengtheners or weakeners\\

Examples:\\
Answer: ``Mount Everest'' | Reference: ``Everest'' → YES;\\
Answer: ``Albert Einstein'' | Reference: ``Einstein'' → YES;\\
Answer: ``Paris'' | Reference: ``The capital of France'' → YES;\\
Answer: ``Pacific Ocean'' | Reference: ``Atlantic Ocean'' → NO;\\
Answer: ``Jupiter'' | Reference: ``I think it's Jupiter'' → YES;\\
Answer: ``Mars'' | Reference: ``Maybe Venus'' → NO.\\

Answer only with:\\
1 if semantically equivalent; \\
0 otherwise.\\

Question: \textless question\textgreater \\
Answer: \textless answer\textgreater \\
Reference answer: \textless target\textgreater \\
Equivalent:
\end{tcolorbox}
\caption{Prompt for judging correctness.}\label{prompt:correctness}
\end{table*}

\begin{table*}[hbp]
\centering
\begin{tcolorbox}[
    width=\textwidth,
    colback=white,
    colframe=black,
    title=Semantic Equivalence Judgment Prompt for Semantic Grouping,
    fonttitle=\bfseries,
    boxrule=0.5pt,
    arc=2pt,
    left=4pt,
    right=4pt,
    top=4pt,
    bottom=4pt,
]
\small\ttfamily
You are grouping 10 answers by their semantic equivalence. Your goal is to judge meaning, not exact wording. Two answers should belong to one group if they are semantically equivalent, even if they use different phrasing, synonyms, or slightly different names that refer to the same entity.\\

Also, treat answers as equivalent even if they include expressions of confidence or uncertainty (e.g., 'I'm pretty sure it's X' vs. 'X'). Focus on the core meaning of the answers and ignore differences in:\\
- wording or phrasing\\
- length\\
- fillers\\
- strengtheners or weakeners\\

Examples:\\
- Answer: 'Mount Everest' | Reference: 'Everest' → YES\\
- Answer: 'Albert Einstein' | Reference: 'Einstein' → YES\\
- Answer: 'Paris' | Reference: 'The capital of France' → YES\\
- Answer: 'Pacific Ocean' | Reference: 'Atlantic Ocean' → NO\\
- Answer: 'Jupiter' | Reference: 'I think it's Jupiter' → YES\\
- Answer: 'Mars' | Reference: 'Maybe Venus' → NO\\

Answer in a dict format only, answers in the same group fall into one dict, for example:\\
$\{$1: answer1, 2: ...$\}$\\
$\{$4: answer4, 5: ...$\}$\\

Question: \textless question\textgreater \\
Answer: \textless sampled answers \textgreater 
\end{tcolorbox}
\caption{Prompt for semantic grouping.}\label{prompt:grouping}
\end{table*}

\subsection{Evaluated LLMs} \label[appendix]{app:llm-detail}
We evaluate 11 LLMs spanning five major families, which consist of instruction and reasoning models.  

\begin{itemize}
    \item From \textbf{Mistral}, we use Mistral-123B (\texttt{Mistral-Large-Instruct-2411}) \citep{mistral2024largeinstruct} and Ministral 8B and 14B reasoning models (\texttt{Ministral-3-Reasoning-2512}) \citep{mistralai2025ministral3reasoning8b}.
    \item  From \textbf{Llama}, we test Llama-70B (\texttt{Llama-3.3-70B-Instruct}; \citealp{meta2024llama3}) . 
    
    \item From \textbf{Qwen}, we cover three model sizes: Qwen2.5-72B, Qwen2.5-32B, and Qwen2.5-14B (\texttt{Qwen2.5-Instruct}, \citealp{qwen2024qwen2.5}).

    \item From \textbf{OLMo}, we assess a broad range of models \cite{olmo2024furious}, including \texttt{OLMo-2-0325-32B-Instruct} (OLMo-32B), \texttt{OLMo-2-1124-13B} (OLMo-13B), and \texttt{OLMo-2-1124-7B} (OLMo-7B).

    \item  From \textbf{GPT}, we include GPT-4o (\texttt{gpt-4o-2024-11-20}) \citep{openai2024gpt4o}, as it is a commercial model, confidence estimation methods (such as Boosted Prob. and internal states-based methods) that require beyond the top 20 output logits are not applicable.  
\end{itemize}

\subsection{Implementing Detail}\label[appendix]{implement-detail}

\paragraph{Confidence Estimation Methods.} Below are implementation details of the evaluated CE methods.

\begin{itemize}
    \item \textbf{Seq. Likelihood.} We use the normalized probability of the generated response as a likelihood-based baseline. Specifically, we sum token log-probabilities over generated tokens only, excluding prompt tokens, and divide by the response length. 
    \item \textbf{Boosted Prob.} Boosted probability is computed from generated-token probabilities following prior probability-based uncertainty estimators \citep{dinh-niehues-2025-generative}.  We extract logits only for generated tokens and compute the boosted-probability statistic over the model's output distribution. 
    \item \textbf{Platt Scaling \citep{platt1999probabilistic}. } Platt scaling trains two scalars over sequence likelihood and correctness labels. The input feature is sequence-likelihood, and the scalars learn a one-dimensional affine transformation followed by a sigmoid.   

    \item \textbf{P(True)} \citep{kadavath2022language}. P(True) is implemented as a self-evaluation query to the same language model. We construct a binary verification prompt containing the question and proposed answer, ending with a request to choose whether the proposed answer is ``True'' or ``False''. We then compute the normalized probability of the True option from the model's next-token log probabilities.

    \item  \textbf{Calib1} \citep{xia-etal-2025-influences}. Calib1 is implemented as a supervised text-pair calibrator. We encode the question and answer with a BERT-style encoder and train a classifier to predict answer correctness. The calibrator operates only on the textual question-answer pair and does not require access to internal states of the target LLM. We train the classifier with focal loss \citep{mukhoti2020calibrating} and use the predicted positive-class probability as the confidence score.

    \item \textbf{Verbalized Confidence} \citep{tian2023just}. Verbalized confidence is implemented as a second-stage prompting method. After the model produces an answer, we ask it to explicitly report a confidence value. 

    \item \textbf{Attention Score }\citep{NEURIPS2024_3c1e1fdf}. The attention score is implemented as a white-box statistic over transformer attention. During local inference, we extract attention from a mid-to-late layer, approximately 70\% through the model depth. We immediately reduce the attention tensor to a scalar using an eigenspectrum-based statistic. The scalar attention feature is then mapped to a confidence score with a logistic regression calibrator trained on correctness labels.

    \item \textbf{Hidden Score} \citep{NEURIPS2024_3c1e1fdf}. Similar to the attention score, the hidden score is implemented from the same mid-to-late layer hidden representations. We extract the relevant hidden-state vector and compute a singular-value-based scalar statistic. We then train a logistic regression calibrator over this scalar feature to produce a normalized confidence score.

    \item \textbf{SAPLMA \citep{azaria-mitchell-2023-internal}.} SAPLMA is implemented as a learned probe over the language model's final-layer hidden state. For each generated answer, we use the hidden state associated with the final generated-token position. We train an MLP classifier on these fixed representations to predict correctness. The LLM parameters are frozen throughout training; only the MLP probe is optimized. The confidence score is the classifier's positive-class probability.

    \item \textbf{P(IK) \citep{kadavath2022language}.}
P(IK) is implemented as a pre-generation linear probe. Instead of using a representation from the generated answer, we use the final-layer hidden state corresponding to the question's last token. A single linear layer with sigmoid output is trained to predict whether the model will answer correctly. The language model remains frozen, and only the linear head is trained.
    
\end{itemize}

\end{CJK}
\end{document}